\documentclass[10pt,onecolumn]{article}

\usepackage[margin=0.75in]{geometry}
\usepackage{times}
\usepackage{amsmath}
\usepackage{amssymb}
\usepackage{amsfonts}
\usepackage{booktabs}
\usepackage{array}
\usepackage{longtable}
\usepackage{url}
\usepackage{hyperref}
\usepackage{graphicx}
\usepackage[utf8]{inputenc}
\usepackage{textgreek}
\usepackage{balance}
\usepackage{float}
\usepackage{wrapfig}

\usepackage{threeparttable}

\usepackage{cite}  
\usepackage{listings}
\usepackage{xcolor}
\definecolor{codegreen}{rgb}{0,0.6,0}
\definecolor{codegray}{rgb}{0.5,0.5,0.5}
\definecolor{codepurple}{rgb}{0.58,0,0.82}

\lstdefinelanguage{yaml}{
  keywords={true,false,null,yes,no},
  keywordstyle=\color{codepurple}\bfseries,
  sensitive=false,
  comment=[l]{\#},
  commentstyle=\color{codegreen}\ttfamily,
  stringstyle=\color{codepurple}\ttfamily,
  morestring=[b]',
  morestring=[b]"
}

\usepackage{titling}
\pretitle{\centering\LARGE\bfseries}
\posttitle{\par\vskip 0.5em}

\preauthor{\centering\large}
\postauthor{\par}

\predate{\centering\large}
\postdate{\par\vskip 1em}

\usepackage{titlesec}
\titleformat{\section}{\large\bfseries}{\thesection}{1em}{}
\titleformat{\subsection}{\normalsize\bfseries}{\thesubsection}{1em}{}
\titleformat{\subsubsection}{\normalsize\bfseries}{\thesubsubsection}{1em}{}

\renewenvironment{abstract}
{\small\quotation\noindent\textbf{Abstract}\par}
{\endquotation}

\begin{document}

\title{Towards a Standard, Enterprise-Relevant Agentic AI Benchmark\vspace{0.5em}\\ \large Lessons from 5.5 billion tokens' worth of agentic AI evaluations}

\author{JV Roig\\
\small Kamiwaza AI\\
\small \texttt{jv@kamiwaza.ai}
}

\date{October 2025}

\maketitle

\begin{abstract}
Enterprise adoption of agentic AI systems requires reliable evaluation methods that reflect real-world deployment scenarios. Traditional LLM benchmarks suffer from training data contamination and fail to measure agentic capabilities such as multi-step tool use and decision-making under uncertainty. We present the Kamiwaza Agentic Merit Index (KAMI) v0.1, an enterprise-focused benchmark that addresses both contamination resistance and agentic evaluation. Through 170,000 LLM test items processing over 5.5 billion tokens across 35 model configurations, we demonstrate that traditional benchmark rankings poorly predict practical agentic performance. Notably, newer generation models like Llama 4 or Qwen 3 do not always outperform their older generation variants on enterprise-relevant tasks, contradicting traditional benchmark trends. We also present insights on cost-performance tradeoffs, model-specific behavioral patterns, and the impact of reasoning capabilities on token efficiency---findings critical for enterprises making deployment decisions.
\end{abstract}


\section{Introduction}
Evaluating the real-world effective capability of LLMs is a continuing problem for the enterprise. This leads to mismatched expectations for prototyping and deployment that cause failures or aborted, failed pilots. 

Traditional benchmark scores don’t  translate to real-world performance for various use-cases and are no better guides in this matter. Worse, they can mislead agentic AI implementation developers into focusing on the wrong models based on various benchmark scores that do not translate to real-world performance. Most benchmarks also fail to capture the reality of enterprise agentic AI - very narrow, specialized, repetitive tasks that are mostly mundane and boring that can be handled either by a small model, or a group of smaller models each for dedicated tasks and use cases. \cite{belcak2025smalllanguagemodelsfuture} 

Having a standard benchmark that is directly useful and relevant to enterprises is a step in the right direction. This would give enterprises a standard benchmark they can reasonably base agentic AI-related decisions on, such as which big or small LLM is up to the task they have in mind, be it from the overall benchmark scores, or specific subscores they are interested in. This in turn could help estimate realistic costs better, increasing project feasibility and the chances of success. Towards this goal, Kamiwaza AI researchers have explored how to create such a benchmark - the Kamiwaza Agentic Merit Index (KAMI).

From the process of experimentation towards this goal, we release not only the alpha (v0.1) version results across a few popular models of different sizes, but also the interesting insights we’ve gained from this experience.

In spirit, our goal with KAMI parallels what the SPEC CPU benchmark suite achieved for computer architecture research and enterprise hardware evaluation~\cite{speccpu}. Before SPEC CPU, processor performance was typically reported through synthetic or vendor-specific microbenchmarks that bore little resemblance to real workloads. SPEC CPU standardized evaluation around realistic, reproducible software tasks like compilation, compression, and numerical computation, providing a neutral, trusted reference for enterprise decision-making. In an analogous way, KAMI aims to become a standardized, contamination-resistant measure of real-world, enterprise-relevant agentic AI capability, replacing narrow, academic leaderboard metrics with assessments that better capture practical deployment behavior.

\section{Background and Related Work}

This section provides context for KAMI v0.1 development and situates our work within the broader landscape of LLM evaluation research. We first examine the fundamental limitations of existing LLM benchmarks that motivated our approach. We then review the extensive literature on benchmark contamination and construct validity issues, demonstrating why traditional static benchmarks cannot reliably assess real-world agentic capabilities. Finally, we discuss prior work on agentic AI evaluation that we use as basis for KAMI v0.1.

\subsection{LLM Benchmarking Issues}
\label{sec:llmbenchmarkingissues}
LLM benchmarking is fraught with issues. In general, we identify two general categories of LLM benchmarking issues that affect how typical benchmarks fail the needs of enterprise AI developers:

\textbf{Benchmark data contamination / overfitting to the test:} When benchmark test data is included in training data, be it accidentally or otherwise, this makes LLM performance better on the benchmark without necessarily improving the LLM's actual abilities. The LLM, in essence, overfits to the test during training. This boosts benchmark scores without actually boosting real-world performance, and makes traditional benchmarks a poor reference for choosing LLMs for deployment purposes. As we'll see later on in the paper, the next generation of models are not always superior to the older generation---an observation we've seen apply to at least two popular model families, Llama and Qwen---despite traditional benchmark scores generally showing they do.

\textbf{Agentic disconnect:} This is terminology we use to describe how the things being measured by standard benchmarks are completely different from what enterprises actually need: multi-step tool use, decision-making under uncertainty, real-world task completion, and scenarios that are as close as possible to what enterprises needs---for example, universally-common tasks like CSV analysis and processing, database understanding and processing, etc. Tool use benchmarks like TAU2 \cite{barres2025tau2} and BFCLv3 \cite{patil2025bfcl} mitigate the multi-step and tool-use part, but are still removed from typical agentic usage. As we'll see later on in the paper, the scores in traditional benchmarks don't align very well when compared against scores for real-world, mundane tasks. Most benchmarks also ignore the stochastic nature of LLMs and do not capture what the real-world reliability would be. 

\subsection{Evidence of Contamination in Literature}
\label{sec:contamination-evidence}

Benchmark data contamination has been extensively documented in literature, with far-reaching implications for evaluation validity. Recent interdisciplinary reviews have identified contamination as one of the most critical challenges facing AI evaluation \cite{eriksson2025trustaibenchmarks}.

\subsubsection{Contamination Mechanisms and Evidence}

Data contamination occurs when benchmark test data appears in training datasets, either intentionally or unintentionally, severely compromising test integrity \cite{xu2024benchmark, zhang2024language, magar2022data, roberts2023data}. This manifests through several mechanisms including data leakage \cite{kaufman2012leakage, xu2024benchmarking} and train-test overlap \cite{lewis2021question}, producing effects similar to overfitting and memorization \cite{tirumala2022memorization, magar2022data}. Models affected by contamination exhibit high performance on familiar in-distribution tasks while failing on similar difficulty tasks with distribution shifts \cite{yuan2023revisiting, xu2024benchmark, zhang2024language, narayanan2023gpt4}.

A striking demonstration of this phenomenon comes from Narayanan and Kapoor's evaluation of GPT-4 on Codeforces programming challenges \cite{narayanan2023gpt4}. When tested on problems added to Codeforces before September 5, 2021, GPT-4 could regularly solve ``easy'' difficulty problems. However, for problems added after that date---well within GPT-4's capability range---the model could not solve a single question correctly. This stark performance cliff strongly suggests memorization of questions and answers rather than genuine problem-solving capability. 

Similar patterns have been systematically documented in mathematical reasoning benchmarks. GSM-Symbolic \cite{gsmsymbolic} introduced symbolic templates that generate diverse instantiations of the same underlying question, revealing that LLMs exhibit noticeable variance when only numerical values are altered. Even more dramatically, adding a single seemingly-relevant but ultimately irrelevant clause to questions caused performance drops up to 65\% across all state-of-the-art models tested. RV-Bench \cite{rvbench} further confirms this pattern through random variable questions (RVQs) that mirror original benchmark problems but with randomized variable combinations. The study found consistent ``proficiency imbalance between encountered and `unseen' data distributions,'' with limited proficiency generalization across similar mathematical reasoning tasks.

Despite widespread awareness of contamination issues and available mitigation strategies, reporting remains inadequate. Zhang et al analyzed 30 prominent language models in October 2024 and found that only 9 reported train-test overlap information, leaving the contamination status of the majority unclear \cite{zhang2024language}.

\subsubsection{Construct Validity: Measuring the Wrong Thing}

Beyond contamination, many benchmarks suffer from weak construct validity---they fail to measure what they claim to measure \cite{raji2021ai}. This problem is particularly acute when benchmarks present themselves as measuring universal or general capabilities. Raji et al argue that framing an AI benchmark as general purpose ``is ultimately dangerous and deceptive, resulting in misguidance on task design and focus, underreporting of the many biases and subjective interpretations inherent in the data as well as enabling, through false presentations of performance, potential model misuse''~\cite{raji2021ai}

A related issue is the use of inadequate proxies for complex real-world capabilities. For example, the HellaSwag benchmark uses Amazon crowdworkers and Reddit's ``Am I the asshole?'' forum posts as proxies for ethical and moral reasoning \cite{keegan2024everyone}. Professional certification exams present similar problems. As Narayanan and Kapoor observe, these benchmarks ``emphasize the wrong thing'' and ``overemphasize precisely the thing that language models are good at,'' noting pointedly: ``It's not like a lawyer's job is to answer bar exam questions all day.''~\cite{narayanan2023gpt4}

These construct validity issues directly parallel the ``agentic disconnect'' we identified in Section~\ref{sec:llmbenchmarkingissues}. Just as benchmarks may measure test-taking ability rather than professional competence, they similarly fail to capture the multi-step tool use, decision-making under uncertainty, and real-world task completion that characterize actual enterprise agentic AI deployments.

\subsubsection{Implications for Evaluation Practice}

These documented issues of contamination and weak construct validity collectively demonstrate that static benchmarks with fixed question sets cannot provide reliable capability assessment, especially for typical enterprise usage. Traditional benchmarks measure an ill-defined mixture of genuine capability, memorized patterns, test-taking optimization, and strategic performance management. For enterprises making deployment decisions, this creates unacceptable uncertainty about actual system capabilities in production environments.

\subsection{The PICARD Framework}

We build on top of the PICARD framework \cite{roig2025picard} as the basis of our efforts to build the universal standard for enterprise agentic AI benchmarking.

Earlier work in PICARD presented a conceptual framework that provides a repeatable and scalable way to measure LLM agentic AI performance that not only measures performance in an agentic manner (multi-step, tool use, real-world conditions), but also solves the benchmark contamination problem through randomization of variables in questions and randomizing the sandbox itself (files and data) in such a manner that, given appropriate test design, no LLM will ever see the exact same test twice, throwing memorization (and thus, contamination) out the window.

\section{Kamiwaza Agentic Merit Index}

The Kamiwaza Agent Merit Index v0.1 is the beginning of the effort to create a standard, enterprise-relevant agentic AI benchmark. In this section, we discuss the design principles, test components, metrics, and infrastructure and scale of the benchmarking experiments that went into KAMI v0.1.

\subsection{Design Principles}
We designed our Kamiwaza Agentic Merit Index benchmark with all of these LLM benchmarking issues in mind:

\begin{itemize}
    \item We use the PICARD framework to combat benchmark data contamination and enable true agentic assessment through multi-step tool use and realistic task environments.
    \item We model use cases common to enterprise agentic AI deployments: filesystem operations, text file processing, CSV data analysis, and database querying.
    \item LLM behavior can be notoriously stochastic, so agentic fitness must demonstrate run-to-run reliability in addition to overall accuracy. For this purpose, we collect additional metrics useful for describing the expected reliability of a model. These are described in Section~\ref{sec:kami-metrics-collected}.
\end{itemize}

Much like the SPEC CPU benchmark established a trusted, real-world baseline for evaluating processor performance beyond synthetic microbenchmarks, KAMI aspires to serve a parallel role for enterprise agentic AI systems. It provides a standardized, reproducible way to measure end-to-end capability - bridging the gap between leaderboard performance and real-world reliability in deployed environments.

These principles operationalize our goal of creating a ``SPEC CPU for agentic AI''—a standardized, reproducible benchmark that measures real-world capability rather than leaderboard performance.

\subsection{Task Categories}
\label{sec:task-categories}

In the grand roadmap of KAMI, we plan to encompass an incredibly wide variety of agentic tasks, including measuring capability in vendor-specific environments - for example proficiency in solving RDBMS-related and administrative tasks in Oracle vs PostgreSQL vs SQL Server, or NoSQL proficiency in MongoDB vs DynamoDB vs Cassandra. However, building comprehensive benchmarks for such specialized systems requires extensive infrastructure, validation, domain expertise, and extensions to the underlying PICARD framework.

For v0.1, we therefore focus first on establishing a solid foundation with fundamental task categories that are widely applicable, easier to validate, and still sufficiently challenging to differentiate model capabilities:

\begin{itemize}
    \item \textbf{Sanity Check}: Can LLMs follow simple instructions without excessive tool misuse? This category tests whether models exhibit the kind of baseline instruction-following failures observed in earlier work~\cite{roig2025picard}, such as the Nova Pro behavior where models repeatedly invoke tools inappropriately.
    
    \item \textbf{Filesystem Operations}: Basic file and directory manipulation tasks.
    
    \item \textbf{Text Search and Extraction}: Tasks requiring agents to locate and extract specific information within larger text documents.
    
    \item \textbf{CSV Processing}: Data manipulation tasks involving extracting key business information from CSV files.
    
    \item \textbf{Database Processing (Standard)}: SQL-based tasks using SQLite to extract key business information from a database.
    
    \item \textbf{Database Processing (Easy/Guided)}: Similar database tasks, made slightly easier with more obvious column naming and process hints. 
    
    \item \textbf{Response Format Instruction Following}: The easiest database task, repeated three more times in three separate questions, each one asking the agent to package the final answer differently (e.g., raw text or JSON-only response)
\end{itemize}

These categories represent fundamental capabilities that are likely to generalize reasonably well across specialized domains. 

Succeeding incremental versions of KAMI will gradually include more task categories and vendor-specific environments.

\subsection{The Test Suite}
\label{sec:test-suite}

KAMI v0.1 is composed of 19 PICARD question templates, each instantiated 30 times with independently randomized values (\textit{samples=30}), yielding a total of 570 test items under the 7 different task categories described earlier in Section~\ref{sec:task-categories}. As a brief recap, PICARD question templates can contain variables and/or sandbox data randomization. Depending on the test design, this randomization can either be trivial, or can be significant enough that the challenge rating of a question may drastically change from one sample (independent instantiation) to another.

Further, to capture performance consistency, the test suite is run independently 8 times for each LLM, resulting in a total of 4,560 test items (with some exceptions noted in Section~\ref{sec:models-tested}). Some chosen LLMs were also tested more than once on purpose, on different platforms (Gaudi 3 and MI300X) purely as researcher curiosity to see if platform-level differences can be observed.

The distribution of templates across task categories is shown in Table~\ref{tab:template-distribution}. The complete PICARD test definition YAML for all the 19 question templates is provided in Appendix~\ref{app:test-suite}. In general, for each category, there is a question template designed to be easier (for example, a task with a single goal, such as finding and calculating a desired value like the total sales amount in the customer database), and a variant that is designed to be harder (for example, a task with multiple goals, such as being asked to find and compute several values from a database or set of CSV files).

\begin{table}[ht]
\centering
\caption{Distribution of question templates across task categories}
\label{tab:template-distribution}
\begin{tabular}{lc}
\toprule
\textbf{Task Category} & \textbf{Templates} \\
\midrule
Sanity Check & 2 \\
Filesystem Operations & 2 \\
Text Search and Extraction & 4 \\
CSV Processing & 3 \\
Database Processing (Standard) & 3 \\
Database Processing (Easy/Guided) & 2 \\
Response Format Instruction Following & 3 \\
\midrule
\textbf{Total} & \textbf{19} \\
\bottomrule
\end{tabular}
\end{table}

\subsection{Models Tested}
\label{sec:models-tested}

We tested 35 unique model configurations across three different platforms: AMD MI300X, Intel Gaudi 3, and Anthropic's API. The majority of models (27 configurations) were tested on AMD MI300X servers since they formed the bulk of our infrastructure, with a smaller subset on Intel Gaudi 3 (7 configurations) and one commercial API model (Claude 3.5 Haiku). 

We included models from several prominent families:

\begin{itemize}
    \item \textbf{Qwen family}: Multiple versions including Qwen2.5 (7B, 14B, 32B, 72B), Qwen3 (4B, 4B-Instruct-2507, 8B, 14B, 30B-A3B, 30B-A3B-Instruct-2507, 32B, 235B-A22B, 235B-A22B-Instruct-2507), plus hybrid reasoning enabled for some Qwen3 models (4B, 8B, 14B, 30B-A3B, 32B)  
    \item \textbf{Llama family}: Llama 3.1 (8B, 70B) and Llama 3.3 (70B), plus experimental Llama 4 variants (Scout and Maverick 17B)
    \item \textbf{Claude}: Claude 3.5 Haiku (via Anthropic API)
    \item \textbf{Mistral}: Mistral Large Instruct 2411
    \item \textbf{Phi}: Phi-4
\end{itemize}

The Qwen models are particularly well suited for analysis due to the wide spectrum of model sizes available, enabling studies of performance scaling across parameter sizes.

Due to time constraints for the v0.1 milestone, we were unable to include several notable models:

\begin{itemize}
    \item \textbf{Llama 3.1 405B}: Slow inference. Each model needs to answer 4,560 questions total across all 8 runs, most of which require multi-step inferences. Assuming each question takes an average of 6 inferences to solve, this is almost 30,000 inferences, so big and slow models take an inordinate amount of time and introduce heavy opportunity costs (i.e., being able to examine the behavior and failure modes of a variety of smaller models instead)
    \item \textbf{DeepSeek}: Slow inference, plus very long outputs for the reasoning models
    \item \textbf{GPT-OSS}: Harmony format---our existing infrastructure needed tweaking that did not make it in time for v0.1
    \item \textbf{GLM-4}: The exciting GLM-4.6 model was released while our experiments were already wrapping up
\end{itemize}

We aim to include all of these models (and more) in succeeding versions of KAMI. More proprietary models are also planned for inclusion.

Some of these models also had FP8 variants tested (with surprising results). More quantized variants are also planned for inclusioin in future KAMI versions.

\textbf{Test run variations:} While most models completed 8 independent runs of the full 570-item test suite, some exceptions exist due to some eval infrastructure problems we encountered (some insights here are shared in Section~\ref{sec:insights}). Specifically:
\begin{itemize}
    \item \textbf{8 runs} (standard): 29 model configurations
    \item \textbf{7 runs}: qwen2\_5\_72b\_instruct, phi\_4
    \item \textbf{6 runs}: llama\_3\_1\_8b\_instruct
    \item \textbf{5 runs}: qwen3\_8b (non-reasoning)
    \item \textbf{3 runs}: claude\_3\_5\_haiku\_20241022, qwen3\_4b (non-reasoning)
\end{itemize}

\subsection{Metrics Collected}
\label{sec:kami-metrics-collected}

To assess the performance and reliability of LLMs in agentic tasks, we compute several statistical measures across multiple independent test runs.

\subsubsection{Pooled Accuracy}

The pooled accuracy represents the proportion of all test items answered correctly across all runs:

\begin{equation}
\hat{p}_{\text{pooled}} = \frac{\sum_{i=1}^{R} C_i}{\sum_{i=1}^{R} T_i}
\label{eq:pooled-accuracy}
\end{equation}

where $C_i$ is the number of correct responses in run $i$, $T_i$ is the total number of trials in run $i$, and $R$ is the number of runs.

Since all runs in our evaluation use the same 19-question test suite ($T_i = 19$ for all $i$), the pooled accuracy is mathematically equivalent to the mean per-run accuracy. We report pooled accuracy as it naturally extends to scenarios with varying trial counts and provides a clearer interpretation as the overall proportion of correct responses.

\subsubsection{Standard Deviation Across Runs}

The standard deviation measures the variability of model performance across independent test sessions:

\begin{equation}
s = \sqrt{\frac{1}{R - 1} \sum_{i=1}^{R} (a_i - \bar{a})^2}
\label{eq:std-dev}
\end{equation}

where $a_i$ is the accuracy of run $i$, and $\bar{a}$ is the mean accuracy across runs.

This metric quantifies consistency: low standard deviation indicates stable behavior, while high standard deviation suggests fragility or non-determinism. In enterprise settings where reliability is as important as average performance, understanding this variability is crucial for deployment decisions.

\subsubsection{Relative Standard Error of Standard Deviation (RSE)}

The RSE quantifies the relative uncertainty in our estimated standard deviation due to finite replication:

\begin{equation}
\text{RSE}(s) \approx \frac{1}{\sqrt{2(R - 1)}}
\label{eq:rse}
\end{equation}

With only 8 runs (our standard), RSE is approximately 26.7\%, meaning the standard deviation estimate itself could vary by about 27\% due to sampling variation. This helps contextualize stability claims---an apparent difference in variability between two models may not be statistically meaningful.

\subsubsection{95\% t-Based Confidence Interval}

We compute a 95\% confidence interval for the mean accuracy using the t-distribution:

\begin{equation}
\bar{a} \pm t_{(0.975,\, R-1)} \cdot \frac{s}{\sqrt{R}}
\label{eq:t-ci}
\end{equation}

where $\bar{a}$ is the mean accuracy, $s$ is the standard deviation, and $t_{(0.975,\, R-1)}$ is the critical value from the t-distribution with $R-1$ degrees of freedom.

For small sample sizes, normal approximation underestimates uncertainty. The t-based confidence interval is appropriately wider, providing a more honest assessment of the uncertainty in estimated performance given our limited replication.

\subsubsection{Additional Metrics}

Beyond accuracy statistics, we also track:
\begin{itemize}
    \item \textbf{Average Time per Conversation}: Wall-clock time in seconds for the model to complete each test item, including all tool calls, the processing time (CPU, I/O) of the underlying tool operations, and reasoning steps
    \item \textbf{Average Tokens per Conversation}: Total tokens generated (output tokens only) per test item, relevant to understanding computational cost and efficiency trade-offs
\end{itemize}

These auxiliary metrics enable analysis of the accuracy-efficiency trade-off, particularly important when comparing standard instruction-tuned models against reasoning-focused variants.

\subsubsection{Wall-Time Comparison Limitations} 
\label{sec:wall-time-limitations}
Wall-time figures are informational only and should be compared across model variants within the same platform, not across platforms.

\textbf{We did not optimize any of our models for inference speed on any of our servers.} Our configuration is minimally what it takes to make the models run, mostly with tensor parallelism set to 8 or whatever the maximum setting the specific model allows. Even the vLLM version is not kept consistent across different models---even within a platform---because we found some models require a lower or higher version of vLLM to launch successfully.

Our I/O infrastructure also differs significantly between our two inference platforms. Our AMD servers use a single NVMe drive for the experiment, while our Gaudi system uses a local multi-NVMe RAID0 array. Other specifications (memory speed, CPU) were also not standardized across platforms, so no apples-to-apples comparison between MI300X and Gaudi 3 inference speed is possible. Since our wall-time measurements include agentic tool use that reads and writes to files and databases, system-level differences beyond the inference accelerators can contribute a meaningful amount in wall-time.

Therefore, wall-time comparisons are meaningful only for understanding relative performance of different models on the same hardware platform under the same configuration, not for making absolute performance claims or cross-platform comparisons.

\subsection{Experimental Infrastructure and Scale}

This evaluation was conducted across substantial compute resources consisting of 32 AMD MI300X GPUs (distributed across four servers) and 8 Intel Gaudi 3 accelerators (one server), with additional testing via Anthropic's commercial API. 

\begin{table}[ht]
\centering
\caption{Evaluation Scale by Platform}
\label{tab:eval_scale}
\begin{tabular}{lrrrrr}
\hline
\textbf{Platform} & \textbf{Conversations} & \textbf{Input Tokens} & \textbf{Output Tokens} & \textbf{Total Tokens} & \textbf{Wall Time (hrs)} \\
\hline
AMD MI300X Server 1 & 27,633 & 846.7M & 85.3M & 932.0M & 802.6 \\
AMD MI300X Server 2 & 12,877 & 365.5M & 6.6M & 372.1M & 129.6 \\
AMD MI300X Server 3 & 31,888 & 1,126.0M & 59.3M & 1,185.3M & 491.8 \\
AMD MI300X Server 4 & 58,665 & 1,687.1M & 135.5M & 1,822.7M & 1,382.5 \\
Intel Gaudi3 & 37,754 & 1,083.0M & 45.3M & 1,128.3M & 701.7 \\
Anthropic API & 1,710 & 73.9M & 3.4M & 77.3M & 33.1 \\
\hline
\textbf{TOTAL} & \textbf{170,527} & \textbf{5,182.2M} & \textbf{335.5M} & \textbf{5,517.7M} & \textbf{3,541.3} \\
\hline
\end{tabular}
\end{table}

The evaluation had 35 distinct model tests (including the same models tested across different hardware platforms to assess platform-specific differences), each one with 19 question types, 30 independent samples per question (PICARD randomization) and repeated over 8 complete test runs. In aggregate, this evaluation processed over 5.5 billion tokens across 170,527 individual test conversations (this includes tokens from failed runs and re-running failed model tests due to various test infrastructure issues encountered) . Total computation time exceeded 3,500 hours (approximately 147 days of wall-clock time), compressed to under two weeks through parallel execution across multiple hardware platforms.

Server-level statistics reflect testing of diverse model architectures (4B--400B parameters, dense and sparse, with and without reasoning capabilities) across different hardware platforms. These statistics illustrate the scale of evaluation and include platform-specific variations, as some models were intentionally tested on both Intel Gaudi 3 and AMD MI300X hardware to identify any platform-dependent behavioral differences.

\section{Results: Agentic Disconnect Illustrated}
\label{sec:results}
We present the overall KAMI v0.1 results in this section, and illustrate agentic disconnect through comparisons with some existing benchmarks.

\subsection{Overall Results}

\begin{table}[ht]
\centering
\caption{KAMI v0.1 Benchmark Results: Model Performance on Enterprise Agentic Tasks}
\label{tab:main-results}
\begin{threeparttable}
\footnotesize
\begin{tabular}{lp{4cm}rrrrrrrr}
\toprule
\textbf{Platform} & \textbf{Model} & \textbf{Runs} & \textbf{Pooled Acc.} & \textbf{Std Dev} & \textbf{RSE} & \textbf{95\% t-CI} & \textbf{Range} \\
\midrule
AMD MI300X & qwen3\_235b\_a22b\_instruct\_2507\_fp8 & 8 & 88.8\% & ±1.19\% & ±26.7\% & (87.8\%, 89.7\%) & 3.33\% \\
Intel Gaudi3 & qwen3\_235b\_a22b\_instruct\_2507 & 8 & 88.4\% & ±1.43\% & ±26.7\% & (87.2\%, 89.6\%) & 3.51\% \\
AMD MI300X & qwen3\_235b\_a22b\_instruct\_2507 & 8 & 88.2\% & ±1.54\% & ±26.7\% & (87.0\%, 89.5\%) & 4.21\% \\
Anthropic & claude\_3\_5\_haiku\_20241022 & 3 & 75.8\% & ±1.36\% & ±50.0\% & (72.5\%, 79.2\%) & 2.63\% \\
AMD MI300X & llama\_4\_maverick\_17b\_128e\_instruct & 8 & 74.6\% & ±0.93\% & ±26.7\% & (73.8\%, 75.3\%) & 2.81\% \\
AMD MI300X & llama\_3\_3\_70b\_instruct\_fp8\_kv & 8 & 74.5\% & ±1.61\% & ±26.7\% & (73.2\%, 75.9\%) & 5.26\% \\
AMD MI300X & llama\_3\_1\_70b\_instruct & 8 & 73.4\% & ±1.46\% & ±26.7\% & (72.2\%, 74.7\%) & 4.74\% \\
AMD MI300X & llama\_4\_maverick\_17b\_128e\_instruct\_fp8 & 8 & 73.1\% & ±1.60\% & ±26.7\% & (71.8\%, 74.5\%) & 4.74\% \\
AMD MI300X & qwen3\_30b\_a3b\_thinkmode & 8 & 72.7\% & ±1.40\% & ±26.7\% & (71.6\%, 73.9\%) & 3.86\% \\
AMD MI300X & llama\_3\_3\_70b\_instruct & 8 & 71.6\% & ±1.51\% & ±26.7\% & (70.4\%, 72.9\%) & 5.44\% \\
AMD MI300X & qwen2\_5\_72b\_instruct & 7 & 71.1\% & ±0.78\% & ±28.9\% & (70.4\%, 71.8\%) & 2.46\% \\
AMD MI300X & qwen3\_30b\_a3b\_instruct\_2507 & 8 & 69.6\% & ±0.80\% & ±26.7\% & (69.0\%, 70.3\%) & 2.63\% \\
Intel Gaudi3 & qwen3\_30b\_a3b\_instruct\_2507 & 8 & 69.3\% & ±0.89\% & ±26.7\% & (68.6\%, 70.1\%) & 2.81\% \\
AMD MI300X & qwen3\_14b\_thinkmode & 8 & 69.1\% & ±1.54\% & ±26.7\% & (67.8\%, 70.4\%) & 4.39\% \\
Intel Gaudi3 & qwen3\_235b\_a22b & 8 & 67.7\% & ±1.17\% & ±26.7\% & (66.7\%, 68.6\%) & 3.33\% \\
AMD MI300X & qwen3\_32b\_thinkmode & 8 & 67.6\% & ±2.13\% & ±26.7\% & (65.8\%, 69.4\%) & 6.32\% \\
AMD MI300X & qwen2\_5\_14b\_instruct & 8 & 66.6\% & ±1.19\% & ±26.7\% & (65.6\%, 67.6\%) & 3.16\% \\
AMD MI300X & llama\_4\_scout\_17b\_16e\_instruct & 8 & 64.1\% & ±1.66\% & ±26.7\% & (62.7\%, 65.4\%) & 4.91\% \\
AMD MI300X & qwen3\_32b\_fp8 & 8 & 63.7\% & ±0.86\% & ±26.7\% & (63.0\%, 64.4\%) & 2.63\% \\
AMD MI300X & qwen3\_8b\_thinkmode & 8 & 62.5\% & ±1.81\% & ±26.7\% & (61.0\%, 64.1\%) & 5.09\% \\
AMD MI300X & qwen3\_32b & 8 & 61.6\% & ±0.89\% & ±26.7\% & (60.8\%, 62.3\%) & 2.81\% \\
AMD MI300X & qwen3\_14b\_fp8 & 8 & 60.0\% & ±0.66\% & ±26.7\% & (59.5\%, 60.5\%) & 1.75\% \\
Intel Gaudi3 & qwen3\_4b\_instruct\_2507 & 8 & 60.0\% & ±1.04\% & ±26.7\% & (59.1\%, 60.9\%) & 2.81\% \\
AMD MI300X & qwen3\_4b\_instruct\_2507 & 8 & 59.9\% & ±1.01\% & ±26.7\% & (59.1\%, 60.8\%) & 3.16\% \\
Intel Gaudi3 & qwen3\_32b & 8 & 59.7\% & ±1.32\% & ±26.7\% & (58.6\%, 60.8\%) & 3.33\% \\
Intel Gaudi3 & mistral\_large\_instruct\_2411 & 8 & 58.9\% & ±1.58\% & ±26.7\% & (57.6\%, 60.2\%) & 4.56\% \\
AMD MI300X & qwen3\_14b & 8 & 58.7\% & ±0.68\% & ±26.7\% & (58.2\%, 59.3\%) & 1.75\% \\
AMD MI300X & qwen3\_30b\_a3b & 8 & 58.1\% & ±0.88\% & ±26.7\% & (57.4\%, 58.8\%) & 2.28\% \\
AMD MI300X & qwen2\_5\_32b\_instruct & 8 & 55.9\% & ±1.29\% & ±26.7\% & (54.8\%, 57.0\%) & 3.86\% \\
Intel Gaudi3 & phi\_4 & 7 & 54.8\% & ±1.30\% & ±28.9\% & (53.6\%, 56.0\%) & 3.86\% \\
AMD MI300X & qwen3\_4b\_thinkmode & 8 & 50.5\% & ±1.55\% & ±26.7\% & (49.2\%, 51.8\%) & 4.21\% \\
AMD MI300X & qwen3\_8b & 5 & 49.1\% & ±0.99\% & ±35.4\% & (47.8\%, 50.3\%) & 2.63\% \\
AMD MI300X & qwen2\_5\_7b\_instruct & 8 & 41.6\% & ±1.21\% & ±26.7\% & (40.5\%, 42.6\%) & 3.68\% \\
AMD MI300X & qwen3\_4b & 3 & 37.8\% & ±1.41\% & ±50.0\% & (34.3\%, 41.3\%) & 2.81\% \\
AMD MI300X & llama\_3\_1\_8b\_instruct & 6 & 10.5\% & ±0.95\% & ±31.6\% & (9.5\%, 11.5\%) & 2.46\% \\
\bottomrule
\end{tabular}
\begin{tablenotes}
\footnotesize
\item Models sorted by mean accuracy. RSE (Relative Standard Error) reflects uncertainty in std dev estimates. t-CI is the 95\% confidence interval for mean accuracy using t-distribution (appropriate for small sample sizes). (see Section~\ref{sec:kami-metrics-collected}).
\end{tablenotes}
\end{threeparttable}
\end{table}

Table~\ref{tab:main-results} shows the main KAMI v0.1 benchmark results across 35 model configurations. We expected most models to score reliably high on these fundamental agentic tasks. Surprisingly, several models we anticipated would be more proficient ended up performing worse than expected (e.g., the original Qwen3 compared to the older Qwen2.5), revealing a disconnect between general capability benchmarks and enterprise agentic performance. We discuss some of these surprises in the following sections.

\textbf{Interpreting Qwen3 model names:} In Table~\ref{tab:main-results}, we observe the following naming conventions for the various Qwen3 models:

\begin{itemize}
    \item \textbf{[Qwen3 Name only]}: The original Qwen3 model release, reasoning disabled. For example, \texttt{qwen3\_30b\_a3b}
    \item \textbf{[Qwen3 Name + thinkmode]}: The original Qwen3 model release, reasoning enabled. For example, \\\texttt{qwen3\_30b\_a3b\_thinkmode}
    \item \textbf{[Qwen3 Name + instruct\_2507]}: The improved Qwen3 model where hybrid thinking was removed. ``Instruct'' denotes a dedicated non-reasoning model. For example, \texttt{qwen3\_30b\_a3b\_instruct\_2507}
\end{itemize}

In addition, models ending with \texttt{\_fp8} denote FP8-quantized variants.

\begin{table}[ht]
\centering
\caption{KAMI v0.1 Benchmark Results - Timing and tokens per conversation}
\label{tab:timing-results}
\begin{threeparttable}
\footnotesize
\begin{tabular}{lp{4cm}rrrr}
\toprule
\textbf{Platform} & \textbf{Model} & \textbf{Pooled Acc.} & \textbf{95\% t-CI} & \textbf{Avg Time (s)} & \textbf{Avg Tokens} \\
\midrule
AMD MI300X & qwen3\_235b\_a22b\_instruct\_2507\_fp8 & 88.8\% & (87.8\%, 89.7\%) & 119.12 & 840.9 \\
Intel Gaudi3 & qwen3\_235b\_a22b\_instruct\_2507 & 88.4\% & (87.2\%, 89.6\%) & 72.12 & 818.1 \\
AMD MI300X & qwen3\_235b\_a22b\_instruct\_2507 & 88.2\% & (87.0\%, 89.5\%) & 87.47 & 811.4 \\
Anthropic & claude\_3\_5\_haiku\_20241022 & 75.8\% & (72.5\%, 79.2\%) & 69.63 & 2013.9 \\
AMD MI300X & llama\_4\_maverick\_17b\_128e\_instruct & 74.6\% & (73.8\%, 75.3\%) & 96.56 & 1856.4 \\
AMD MI300X & llama\_3\_3\_70b\_instruct\_fp8\_kv & 74.5\% & (73.2\%, 75.9\%) & 24.95 & 894.6 \\
AMD MI300X & llama\_3\_1\_70b\_instruct & 73.4\% & (72.2\%, 74.7\%) & 38.91 & 1059.6 \\
AMD MI300X & llama\_4\_maverick\_17b\_128e\_instruct\_fp8 & 73.1\% & (71.8\%, 74.5\%) & 91.09 & 1796.1 \\
AMD MI300X & qwen3\_30b\_a3b\_thinkmode & 72.7\% & (71.6\%, 73.9\%) & 211.97 & 5217.3 \\
AMD MI300X & llama\_3\_3\_70b\_instruct & 71.6\% & (70.4\%, 72.9\%) & 25.69 & 739.9 \\
AMD MI300X & qwen2\_5\_72b\_instruct & 71.1\% & (70.4\%, 71.8\%) & 97.02 & 1498.4 \\
AMD MI300X & qwen3\_30b\_a3b\_instruct\_2507 & 69.6\% & (69.0\%, 70.3\%) & 45.99 & 731.4 \\
Intel Gaudi3 & qwen3\_30b\_a3b\_instruct\_2507 & 69.3\% & (68.6\%, 70.1\%) & 33.07 & 726.0 \\
AMD MI300X & qwen3\_14b\_thinkmode & 69.1\% & (67.8\%, 70.4\%) & 118.32 & 4929.5 \\
Intel Gaudi3 & qwen3\_235b\_a22b & 67.7\% & (66.7\%, 68.6\%) & 62.60 & 527.2 \\
AMD MI300X & qwen3\_32b\_thinkmode & 67.6\% & (65.8\%, 69.4\%) & 162.76 & 4734.6 \\
AMD MI300X & qwen2\_5\_14b\_instruct & 66.6\% & (65.6\%, 67.6\%) & 53.97 & 1312.6 \\
AMD MI300X & llama\_4\_scout\_17b\_16e\_instruct & 64.1\% & (62.7\%, 65.4\%) & 29.89 & 1167.6 \\
AMD MI300X & qwen3\_32b\_fp8 & 63.7\% & (63.0\%, 64.4\%) & 46.17 & 490.1 \\
AMD MI300X & qwen3\_8b\_thinkmode & 62.5\% & (61.0\%, 64.1\%) & 150.07 & 6663.0 \\
AMD MI300X & qwen3\_32b & 61.6\% & (60.8\%, 62.3\%) & 30.40 & 467.8 \\
AMD MI300X & qwen3\_14b\_fp8 & 60.0\% & (59.5\%, 60.5\%) & 25.85 & 436.3 \\
Intel Gaudi3 & qwen3\_4b\_instruct\_2507 & 60.0\% & (59.1\%, 60.9\%) & 32.38 & 1184.1 \\
AMD MI300X & qwen3\_4b\_instruct\_2507 & 59.9\% & (59.1\%, 60.8\%) & 40.07 & 884.5 \\
Intel Gaudi3 & qwen3\_32b & 59.7\% & (58.6\%, 60.8\%) & 22.14 & 476.6 \\
Intel Gaudi3 & mistral\_large\_instruct\_2411 & 58.9\% & (57.6\%, 60.2\%) & 43.27 & 610.1 \\
AMD MI300X & qwen3\_14b & 58.7\% & (58.2\%, 59.3\%) & 20.83 & 448.7 \\
AMD MI300X & qwen3\_30b\_a3b & 58.1\% & (57.4\%, 58.8\%) & 31.05 & 448.3 \\
AMD MI300X & qwen2\_5\_32b\_instruct & 55.9\% & (54.8\%, 57.0\%) & 57.45 & 977.2 \\
Intel Gaudi3 & phi\_4 & 54.8\% & (53.6\%, 56.0\%) & 18.27 & 957.3 \\
AMD MI300X & qwen3\_4b\_thinkmode & 50.5\% & (49.2\%, 51.8\%) & 298.71 & 12994.6 \\
AMD MI300X & qwen3\_8b & 49.1\% & (47.8\%, 50.3\%) & 38.77 & 632.0 \\
AMD MI300X & qwen2\_5\_7b\_instruct & 41.6\% & (40.5\%, 42.6\%) & 35.47 & 1012.6 \\
AMD MI300X & qwen3\_4b & 37.8\% & (34.3\%, 41.3\%) & 54.86 & 904.5 \\
AMD MI300X & llama\_3\_1\_8b\_instruct & 10.5\% & (9.5\%, 11.5\%) & 57.27 & 6800.3 \\
\bottomrule
\end{tabular}
\begin{tablenotes}
\footnotesize
\item Models sorted by pooled accuracy. Avg Time measured in seconds per conversation. Avg Tokens represents average output tokens per conversation (see Section~\ref{sec:kami-metrics-collected}).
\end{tablenotes}
\end{threeparttable}
\end{table}

Table \ref{tab:timing-results} shows the KAMI v0.1 with timing information. As discussed in Section \ref{sec:wall-time-limitations}, wall-time here is informational only, and are not apples-to-apples across platforms, as we did not optimize any of our platforms and only did the bare minimum to get models to run reliably, which included mixing vLLM versions and, for ease of batch processing, standardizing tensor parallelism values across models that may not need the high parallelism at all. The slow FP8 inference of \texttt{qwen3\_235b\_a22b\_instruct\_2507\_fp8} might be because of expert parallelism, which we had to enable to get it running at all.

Whether we consider wall-time or avg tokens produced, the reasoning models show their weakness. They do provide significant boosts for our smaller models, but at the cost of much more tokens produced---and consequentially, wall-time. 

\texttt{qwen3\_4b} is a good example. With reasoning disabled, it performs abysmally with a pooled accuracy of 37.8\%, with an average of 900 tokens per conversation. Enabling reasoning boosts this to 50.5\%, but at the cost of 14x more tokens - an average of 13K tokens per conversation.

\subsection{Expected vs Actual LLM Rankings}
\label{sec:benchmark-comparison}

To illustrate the impact of agentic disconnect with existing LLM benchmarks, we compare KAMI v0.1 results with some popular benchmarks, in isolation and in aggregate.

\subsubsection{Against Popular Benchmarks}

One of the biggest and most obvious surprises was how the original Qwen3 models drastically underperformed their older Qwen2.5 counterparts.

\begin{table}[ht]
\centering
\caption{Qwen3 Model Benchmarks from the Qwen Team}
\label{tab:qwen-benchmarks}
\begin{threeparttable}
\footnotesize
\begin{tabular}{lrrrrr}
\toprule
\textbf{Benchmark} & \textbf{Qwen3-235B-A22B} & \textbf{Qwen3-30B-A3B} & \textbf{Qwen3-32B} & \textbf{Qwen3-4B} & \textbf{Qwen2.5-72B-Instruct} \\
\midrule
ArenaHard & 95.6 & 91.0 & 89.5 & 76.6 & 81.2 \\
AIME'24 & 85.7 & 80.4 & 79.5 & 73.8 & 18.9 \\
AIME'25 & 81.5 & 70.9 & 69.5 & 65.6 & 15.0 \\
LiveCodeBench v5 & 70.7 & 62.6 & 62.7 & 54.2 & 30.7 \\
CodeForces & 2056 & 1974 & 1982 & 1671 & 859 \\
Aider (Pass@2) & 61.8 & -- & -- & -- & -- \\
LiveBench (2024-11-25) & 77.1 & 74.3 & 72.0 & 63.6 & 51.4 \\
BFCLv3 & 70.8 & 69.1 & 66.4 & 65.9 & 63.4 \\
MultiIF & 71.9 & 72.2 & 68.3 & 66.3 & 65.3 \\
\bottomrule
\end{tabular}
\begin{tablenotes}
\footnotesize

\item Benchmark results for Qwen3 and Qwen2.5 model families. MoE indicates Mixture of Experts architecture. Dense indicates dense model architecture. Results sourced from Qwen technical reports.
\end{tablenotes}
\end{threeparttable}
\end{table}

Data in Table \ref{tab:qwen-benchmarks} was taken from the Qwen3 launch blog \cite{qwen2025qwen3}. Similar benchmarks showing small Qwen3 models (even 4B) beating their previous open flagship Qwen2.5 72B can be found in the Qwen3 technical report. \cite{yang2025qwen3technicalreport}.

Unfortunately, as we saw in Table~\ref{tab:main-results}, in real agentic tasks, none of the original models - not even the big MoE Qwen3 235B-A22B - was actually better than Qwen2.5 72B. 

This is a prime example of the ``agentic disconnect'' concept we introduced in Section \ref{sec:llmbenchmarkingissues}. 

The various benchmarks found in both the Qwen3 Technical Report \cite{yang2025qwen3technicalreport} and Qwen3 launch blog \cite{qwen2025qwen3} could make a reader reasonably conclude that even the 4B model would be more effective in agentic duty - after all, aside from all the programming and math benchmarks, the new 4B model also outperforms Qwen2.5-72B in BFCLv3!

In reality---as exposed by their performance in KAMI v0.1, where models are expected to do mundane but realistic text, CSV and database data extraction and analysis common in the enterprise---these new small models score far below Qwen2.5-72B.

\begin{table}[ht]
\centering
\caption{$\tau^2$-Bench Telecom (Agentic Tool Use)}
\label{tab:tau2-bench-telecom}
\begin{threeparttable}
\footnotesize
\begin{tabular}{p{4cm}c}
\toprule
\textbf{Model} & \textbf{Score (\%)} \\
\midrule
Qwen2.5 72B & 35 \\
Qwen3 14B (thinking) & 35 \\
Qwen3 235B A22B Instruct 2507 & 33 \\
Qwen3 14B & 32 \\
Mistral Large 2 & 31 \\
Qwen3 32B (thinking) & 30 \\
Qwen3 8B (thinking) & 28 \\
Qwen3 235B & 27 \\
Llama 3.3 70B & 27 \\
Qwen3 4B Instruct 2507 & 27 \\
Qwen3 30B A3B (thinking) & 26 \\
Qwen3 8B & 25 \\
Qwen3 4B (thinking) & 19 \\
Llama 4 Maverick & 18 \\
Llama 3 8B & 16 \\
Llama 4 Scout & 16 \\
Llama 3 .1 70B & 15 \\
Qwen3 30B A3B Instruct 2507 & 10 \\
\bottomrule
\end{tabular}
\begin{tablenotes}
\footnotesize
\item Results sourced from Artificial Analysis.
\end{tablenotes}
\end{threeparttable}
\end{table}

TAU2 bench is another popular benchmark that attempts to measure agentic tool use. We sourced data from Artificial Analysis \cite{artificialanalysis2025} to display TAU2 bench scores the models that we benchmarked for KAMI v0.1, shown in Table \ref{tab:tau2-bench-telecom}. Models in KAMI v0.1 that do not appear here are not available in Artificial Analysis.

The TAU2 bench results shown in Table \ref{tab:tau2-bench-telecom} show a very different reality from the KAMI v0.1. This is the reality of LLM performance varying greatly based on use cases. Models that we have found to be usable in certain SQL, CSV, or text file processing scenarios flounder in TAU2 - for example, Llama 4 Maverick scores incredibly poorly in TAU2 Telecom (18\%), scoring about half as well Qwen3 14B thinking (35\%). 

In KAMI v0.1, we found that both of these models score close enough across all of the filesystem, text, CSV, and database processing tasks, with Maverick having a slight edge. Furthermore, the non-thinking score of Qwen3 14B in TAU2 Telecom (32\%) would also make one expect that it would be superior to Maverick. Instead, we found that Qwen3 14B, without reasoning enabled, fails catastrophically in some CSV, database, and text processing tasks, and score significantly lower overall. Table \ref{tab:kami-maverick-14b-tau} shows this comparison.

\begin{table}[ht]
\centering
\caption{Performance comparison of Llama 4 Maverick and Qwen3 14B (non-thinking and thinking mode) across evaluation runs in filesystem (201, 202), text extraction (301-304), CSV processing (401-403), and Database processing (501-602)}
\label{tab:kami-maverick-14b-tau}
\footnotesize
\begin{tabular*}{\textwidth}{@{\extracolsep{\fill}}lcccccccccccccccc}
\hline
\textbf{Model} & \textbf{q201} & \textbf{q202} & \textbf{q301} & \textbf{q302} & \textbf{q303} & \textbf{q304} & \textbf{q401} & \textbf{q402} & \textbf{q403} & \textbf{q501} & \textbf{q502} & \textbf{q503} & \textbf{q601} & \textbf{q602} \\
\hline
Llama4-Maverick-run1 & 30 & 29 & 29 & 24 & 11 & 2 & 22 & 2 & 21 & 28 & 13 & 19 & 30 & 10 \\
Llama4-Maverick-run2 & 30 & 29 & 29 & 17 & 11 & 4 & 28 & 1 & 21 & 28 & 16 & 20 & 30 & 12 \\
Llama4-Maverick-run3 & 30 & 28 & 30 & 21 & 10 & 4 & 26 & 4 & 20 & 29 & 18 & 18 & 30 & 15 \\
Llama4-Maverick-run4 & 30 & 30 & 29 & 23 & 14 & 1 & 26 & 2 & 18 & 30 & 14 & 17 & 30 & 16 \\
Llama4-Maverick-run5 & 30 & 30 & 30 & 16 & 9 & 5 & 23 & 6 & 17 & 30 & 17 & 17 & 30 & 15  \\
Llama4-Maverick-run6 & 30 & 30 & 29 & 24 & 11 & 5 & 24 & 2 & 22 & 29 & 10 & 18 & 30 & 14 \\
Llama4-Maverick-run7 & 30 & 28 & 30 & 19 & 14 & 1 & 26 & 4 & 23 & 29 & 12 & 10 & 30 & 11 \\
Llama4-Maverick-run8 & 30 & 30 & 27 & 21 & 14 & 3 & 26 & 4 & 19 & 30 & 16 & 12 & 30 & 11 \\
Qwen3-14b-run1 & 30 & 29 & 30 & 27 & 0 & 0 & 0 & 0 & 0 & 10 & 0 & 30 & 30 & 1 \\
Qwen3-14b-run2 & 30 & 30 & 30 & 29 & 0 & 0 & 0 & 0 & 0 & 4 & 0 & 30 & 30 & 2 \\
Qwen3-14b-run3 & 30 & 28 & 30 & 25 & 0 & 0 & 0 & 0 & 0 & 5 & 0 & 30 & 30 & 1 \\
Qwen3-14b-run4 & 30 & 27 & 30 & 27 & 0 & 0 & 0 & 0 & 0 & 4 & 0 & 30 & 30 & 1 \\
Qwen3-14b-run5 & 30 & 29 & 30 & 30 & 0 & 0 & 0 & 0 & 0 & 8 & 1 & 30 & 30 & 1 \\
Qwen3-14b-run6 & 30 & 29 & 30 & 28 & 0 & 0 & 0 & 0 & 0 & 7 & 0 & 30 & 30 & 3 \\
Qwen3-14b-run7 & 30 & 30 & 30 & 30 & 0 & 0 & 1 & 0 & 0 & 4 & 0 & 30 & 30 & 3 \\
Qwen3-14b-run8 & 30 & 28 & 30 & 29 & 0 & 0 & 0 & 0 & 0 & 5 & 0 & 30 & 30 & 3 \\
Qwen3-14b-TM-run1 & 30 & 18 & 30 & 22 & 13 & 7 & 22 & 5 & 11 & 24 & 0 & 29 & 30 & 1 \\
Qwen3-14b-TM-run2 & 29 & 18 & 30 & 17 & 9 & 5 & 21 & 7 & 13 & 25 & 2 & 27 & 30 & 3 \\
Qwen3-14b-TM-run3 & 30 & 21 & 29 & 20 & 13 & 6 & 18 & 5 & 12 & 26 & 0 & 26 & 30 & 0 \\
Qwen3-14b-TM-run4 & 30 & 16 & 30 & 20 & 13 & 3 & 21 & 12 & 18 & 25 & 4 & 28 & 30 & 1 \\
Qwen3-14b-TM-run5 & 30 & 16 & 30 & 19 & 13 & 5 & 15 & 12 & 14 & 28 & 2 & 26 & 30 & 1 \\
Qwen3-14b-TM-run6 & 29 & 20 & 30 & 21 & 13 & 5 & 19 & 8 & 9 & 26 & 2 & 24 & 30 & 1 \\
Qwen3-14b-TM-run7 & 30 & 19 & 30 & 20 & 16 & 6 & 20 & 12 & 17 & 25 & 3 & 30 & 30 & 3 \\
Qwen3-14b-TM-run8 & 30 & 16 & 30 & 20 & 13 & 5 & 18 & 8 & 18 & 26 & 1 & 30 & 30 & 2 \\
\hline
\end{tabular*}
\label{tab:model_results}
\end{table}

The overall rankings in TAU2 Telecom are not predictive of performance in KAMI v0.1. Comparing Tables~\ref{tab:main-results} and~\ref{tab:tau2-bench-telecom}, we observe several striking disconnects:

\begin{itemize}
    \item \textbf{Qwen3 30B-A3B Instruct 2507} ranks at the very bottom of TAU2, yet demonstrates proficient agentic capability in KAMI v0.1 (69.6\%, comparable to Qwen2.5 72B Instruct at 71.1\%).
    
    \item \textbf{Qwen3 235B A22B Instruct 2507} dominates KAMI v0.1 with 88.8\% accuracy, significantly outperforming all other models. In TAU2, it appears merely as part of the upper batch, not even the top performer.
    
    \item \textbf{Llama 4 Scout} ranks near the bottom in TAU2, yet achieves middle-of-the-pack performance in KAMI v0.1 (64.1\%).
    
    \item \textbf{Llama 3.1 70B} also ranks near the bottom in TAU2, but scores among the higher-performing models in KAMI v0.1 (73.4\%), even slightly exceeding Qwen2.5 72B with non-overlapping 95\% confidence intervals.
    
    \item \textbf{Mistral Large 2} ranks 5th out of 18 in TAU2, yet is significantly outperformed in KAMI v0.1 by four of TAU2's bottom five models (the sole exception being Llama 3.1 8B Instruct).
    
    \item \textbf{Llama 3.1 8B Instruct} appears comparable to Llama 3.1 70B, Llama 4 Scout, and Llama 4 Maverick in TAU2---a ranking that seems self-evidently incorrect. In KAMI v0.1, Llama 3.1 8B is the worst-performing model, catastrophically failing nearly all agentic tasks and being the only model in our test set to fail even the basic sanity check tests (q101, q102).
\end{itemize}

More detailed observations could be made, but the central point is not that TAU2 is a poor benchmark in isolation---it is that TAU2, like BFCLv3 before it, fails to serve as a reliable indicator of actual agentic performance in common, mundane enterprise concerns. These benchmarks cannot even be used as rough gauges of agentic fitness for real-world deployment in such routine tasks. This disconnect between benchmark rankings and practical capability illustrates precisely the problem that the KAMI project aims to address: creating evaluation frameworks that are useful not only for academic researchers and model developers, but directly applicable for enterprise deployment decisions.

\subsubsection{Against Aggregated Benchmarks}

If individual benchmarks are bad indicators of useful agentic performance in routine enterprise tasks, could an aggregate of these standard benchmarks perhaps be more indicative, perhaps as a sort of ``wisdom of the crowd'' ?

\begin{table}[ht]
\centering
\caption{Artificial Analysis Intelligence Index v3.0}
\label{tab:aa-intelligence-index}
\begin{threeparttable}
\begin{tabular}{p{6cm}c}
\toprule
\textbf{Model} & \textbf{Index Score} \\
\midrule
Qwen3-235B-A22B Instruct 2507 & 45 \\
Qwen3 32B (thinking) & 39 \\
Qwen3-30B-A3B Instruct 2507 & 37 \\
Qwen3 30B-A3B (thinking) & 37 \\
Qwen3 14B (thinking) & 36 \\
Llama 4 Maverick & 36 \\
Qwen3 4B Instruct 2507 & 30 \\
Qwen3 235B-A22B & 30 \\
Qwen3 14B & 29 \\
Qwen2.5 72B Instruct & 29 \\
Qwen3 8B (thinking) & 28 \\
Llama 4 Scout & 28 \\
Llama 3.3 70B & 28 \\
Mistral Large 2 & 27 \\
Qwen3 30B-A3B & 26* \\
Qwen3 32B & 26* \\
Qwen3 4B (thinking) & 26* \\
Qwen3 8B & 23 \\
Qwen2.5 32B Instruct & 23* \\
Phi 4 & 23 \\
Llama 3.1 70B & 23 \\
Qwen3 4B & 21* \\
Claude 3.5 Haiku & 20* \\
Llama 3.1 8B & 17 \\
\bottomrule
\end{tabular}
\begin{tablenotes}
\footnotesize
\item Intelligence Index v3.0 is composed of 10 evaluations: MMLU-Pro, GPQA Diamond, Humanity's Last Exam, LiveCodeBench, SciCode, AIME 2025, IFBench, AA-LCR, Terminal-Bench Hard, $\tau^2$-Bench Telecom. * indicates estimate.
\end{tablenotes}
\end{threeparttable}
\end{table}

Table \ref{tab:aa-intelligence-index} shows the Artificial Analysis Intelligence Index (AAII) scores of most of the models evaluated in KAMI v0.1 (models that do not appear here are not available in Artificial Analysis). The Intelligence Index is a combination metric composed of 10 different LLM benchmarks: MMLU-Pro, GPQA Diamond, Humanity's Last Exam, LiveCodeBench, SciCode, AIME 2025, IFBench, AA-LCR, Terminal-Bench Hard, and $\tau^2$-Bench Telecom.

Artificial Analysis describes their Intelligence Index as follows~\cite{artificialanalysis2025intelligence}:

\begin{quote}
Artificial Analysis Intelligence Index combines a comprehensive suite of evaluation datasets to assess language model capabilities across reasoning, knowledge, maths and programming. It is a helpful synthesis of overall language model intelligence and can be used to compare language models. Like all evaluation metrics, it has limitations and may not apply directly to every use case. However, we are confident that it is a more useful synthesis comparison between language models than any other metric in existence today.
\end{quote}

However, as we can observe by comparing rankings from Table~\ref{tab:main-results} with the AAII rankings in Table~\ref{tab:aa-intelligence-index}, even this comprehensive aggregation is an unreliable predictor of useful agentic performance in common, mundane enterprise tasks. For example, from the perspective of the common, mundane enterprise tasks that KAMI v0.1 encompasses:

\begin{itemize}
    \item AAII correctly identifies the best performing model (Qwen3-235B-A22B-Instruct-2507) and worst performing model (Llama 3.1 8B Instruct).
    
    \item It drastically underestimates Claude 3.5 Haiku (second worst in AAII, near the top in KAMI v0.1).
    
    \item It drastically underestimates the performance of Llama 3.1 70B (near bottom in AAII, near top in KAMI v0.1).
    
    \item It drastically underestimates the performance of Llama 3.3 70B, where it ranks tied with or lower than models like Scout, Qwen3 4B Instruct 2507, Qwen3 8B (thinking), Qwen3 14B (thinking and non-thinking), Qwen3 235B-A22B (original release, non-thinking), and Qwen3 30B-A3B Instruct 2507 in AAII, while outperforming all of these in KAMI v0.1.
    
    \item It overestimates the performance of Qwen3 4B Instruct 2507 (ranks higher in AAII than many models it underperforms in KAMI v0.1).
    
    \item It correctly captures the relative ranking of Qwen3 14B (non-thinking) and Qwen3 32B (non-thinking), with 14B scoring higher in both benchmarks.
    
    \item There is slight deviation between AAII and KAMI v0.1 for Qwen3 14B (thinking) versus Qwen3 32B (thinking), with AAII slightly favoring Qwen3 32B, while KAMI v0.1 slightly favors Qwen3 14B.
\end{itemize}

Overall, while there are a few correlations found, even the aggregated, wisdom-of-the-crowd approach to using standard LLM benchmarks fails as a useful gauge for agentic performance in common, mundane enterprise tasks. As NVIDIA researchers observe, ``the majority of agentic subtasks in deployed agentic systems are repetitive, scoped, and non-conversational''~\cite{belcak2025smalllanguagemodelsfuture}. The approach of most benchmarks to give a single overall score is an inherent problem: not only does this approach fail to account for these narrow, repetitive tasks that characterize real-world enterprise agentic AI deployments, it also does not provide any information on consistency and reliability.

These are concerns that are baked into the design of KAMI. More enterprise scenarios, including vendor-specific environments, are in the roadmap as mentioned in Section~\ref{sec:task-categories}. KAMI v0.1 already includes metrics on consistency and reliability to distinguish between models that may have the same or similar overall score: run-to-run standard deviation and relative standard error, plus t-CIs to better distinguish whether score differences between two models are real or noise. We also capture additional metrics not yet published in this v0.1 launch, such as Wilson CIs for category-level (instead of test-suite-level) performance comparisons.

\section{Insights}
\label{sec:insights}

In this section, we share some insights derived from our experience with KAMI v0.1 experimentation.

\subsection{Experimental Insights}
\label{sec:experimental-insights}

The KAMI v0.1 benchmarking experience gave us some insights about both the benchmarking infrastructure and the process.

\subsubsection{Benchmarking Infrastructure}

In some cases, we found that during the course of a KAMI test run, a few LLMs eventually do unnecessary, chaotic, hallucinated actions. Llama 3.1 8B Instruct is a prime example. Sometimes, these result in experiment-stopping problems, necessitating a repeat of affected experiment runs. Most of these were due to the generally unrestricted tools available to the LLMs during the experiment, particularly their Python execution tool and filesystem-related tools. For example, we observed a scenario where the agentic server itself was deleted (this is the PICARD-related component that implements the inference loop and tool implementations) and had to be reinstalled. 

In more benign cases, we found the host filesystem littered with new folders and files created in a hallucination-filled agentic adventure.

In another extreme case, entire batches of experiments would suddenly fail at the same time. Root cause analysis revealed that a bug in our agentic server code could get triggered by specific LLM behavior. This causes the agentic server to fail in all succeeding requests. However, we parallelize runs to maximize GPU capacity. Each run has independent PICARD instances, but the agentic server was centralized. This meant that when the bug was triggered, not only would the specific parallel run that triggered it would fail, all of the rest of the parallel runs alongside it would also fail, since they are all served by the same agentic server that is now out of commission. Therefore, aside from fixing the agentic server bug we discovered, a new design decision for kami v0.2 is to have separate agentic server instances when running in parallel to limit the blast radius of unforeseen events. This would have saved us tens of hours in unnecessarily failed runs we had to repeat.

The concern for LLMs wielding all-powerful tools can also be further mitigated by having the agentic server (and thus, the environment the LLM sees and works in) deployed in independent Docker containers. In this way, containers are essentially ephemeral per test and serve as a safer sandbox. If the overhead of running docker containers proves to not be onerous, this would likely be the direction for KAMI v0.2 and onwards.

\subsubsection{Benchmark Efficiency Trade-offs}

During the design phase, we settled on 8 runs with 30 samples for each question template because of the high total trials it produces (240 for each question template) while also having a reasonable trial size per run for per-run SE metrics. In practice, we found this to be too time-consuming and expensive. Finding a lower acceptable setting is one of the next design decisions to be made for v0.2.

In real-world enterprise settings (i.e., using KAMI-like infrastructure to test potential models for specific applications), we recommend a two-stage approach when trying to evaluate a large number of potential models. In the first stage, start with a much smaller total number of trials to separate good candidate models from the bad, such as 3 runs with 20 samples each. In the second stage, after eliminating most of the other models and the goal is to accurately evaluate the top 2 - 3 models to find the most reliable one, do 9 additional runs with 20 samples each. Then, when computing the final metrics for these top models, aggregate the entire 12 runs since sample count was kept constant. In this strategy, we keep the same total trials (240 in our example) and keep a high total run count in order to get tight t-CIs and SDs for the top models, while saving substantial compute resources (or API costs) for filtering out most of the non-ideal models.

The KAMI methodology using the PICARD framework, while framed in the previous example as a way to benchmark models, is also applicable for benchmarking specific prompts and tool designs. Using the same methodology, one would simply keep the model constant while swapping different sets of prompts and/or tools to evaluate which configurations work better.

\subsection{LLM and Agentic AI Insights}
\label{sec:agentic-ai-insights}

Our KAMI v0.1 experience has emphasized some agentic AI insights that affect project success and feasibility. We share some of these insights in this section.

\subsubsection{Bigger is Not Always Better}

A common rule of thumb when thinking about LLM capabilities is ``bigger = better''. While this is true in a general sense, when considering LLM deployment for common, mundane enterprise agentic AI tasks, this is not always the case. Table \ref{tab:q303-performance} shows an example where in a specific KAMI v0.1 question template, the overall champion model (and the second largest model in the current batch) is not the most capable.

\begin{table}[ht]
\centering
\caption{Model Performance on Q303 (Text Search and Extraction, Finding Words in Two Random Positions)}
\label{tab:q303-performance}
\begin{tabular}{lccccl}
\toprule
\textbf{Model} & \textbf{Size} & \textbf{Success Rate} & \textbf{Python Usage} & \textbf{Python Success} & \textbf{Issue} \\
\midrule
Llama 4 Scout & 109B & 80.0\% & 84\% & 90.6\% & Minor execution errors \\
Claude 3.5 Haiku & - & 68.9\% & 76\% & 89--95\% & Semantic confusion \\
Qwen2.5 72b Instruct & 72B & 60.5\% & 95\% & 63.8\% & Poor Python reliability \\
Qwen3 235b Instruct 2507 & 235B & 57.9\% & 59\% & 97.9\% & Random tool selection \\
Qwen3 235b Instruct 2507 FP8 & 235B & 40.0\% & 41\% & 98.0\% & Uses Python less often \\
\bottomrule
\end{tabular}
\end{table}

When we investigated what happened here, we found that essentially Llama 4 Scout won out because it was either confident of its code execution, or simply unconfused by the Python tool success message. Qwen3 235B A22B Instruct 2507, on the other hand, sometimes ended up second-guessing itself if the code execution actually worked or not. As a consequence, Qwen3 235B A22B Instruct 2507 would sometimes conclude the Python tool was not working (it was) and, instead of checking if the necessary file was correctly created (as per the code it wrote), would end up trying to ``eyeball'' or manually count the words without using tools or code. This always fails, as LLMs cannot count objects manually when the positional count is high enough (beyond a few tens, for example).

Claude 3.5 Haiku's failure mode is different. Like Scout, it is either confident of the code it executed or simply unconfused by the Python tool success message. What causes it to get wrong answers is confusing ``words'' with ``lines''. For example, when asked to retrieve the 47th and 169th, it would sometimes try to retrieve the 47th and 169th lines instead.

Regarding the effects of tool messages, we elaborate more on this in Section \ref{sec:tools-are-prompts}.

\subsubsection{Slight Hints Can Drastically Lower the LLM Cost of a Task}

KAMI v0.1 ``Database Processing (Easy/Guided)'' task category presents an easier version of the preceding ``Database Processing (Standard)'' task category, to measure the effect of improved instructions for agentic AI deployments. One of these tasks is \texttt{q601}, the easier version of the \texttt{q501} task. In both of these, the goal is the same: given a database to analyze, retrieve the number of orders beyond a randomized threshold for a randomized region. (The full KAMI v0.1 test suite definition can be found in Appendix \ref{app:test-suite} for the details).

To design \texttt{q601}, we first ran \texttt{q501} on two reasonably-sized LLMs (Qwen2.5 72B and Llama 3.1 70B). Then, we did failure mode analysis to understand how the models usually failed. From there, we modified the question template of \texttt{q501} to fix the common failure modes we observed: guessing the schema (without using the available schema inspection tool), and making assumptions about the values in a necessary column. We advise the LLM to begin by using the schema tool first to retrieve the database schema, and we made the column name more obvious for the column where they were assuming inappropriate values. We could have also solved the column problem by instructing the LLM to sample values from the column first instead of making assumptions, but we were more interested in seeing the effect of the column name change instead of another direct instruction.

The effect was drastic. In \texttt{q501}, the average score per run across all models was 19.56 out of 30. In q601, this increased to a near-perfect 29.25. Only Llama 3.1 8B Instruct still failed the test. The next weakest model, Qwen3 4B (non-thinking), increased its score from an average of 1/30 in \texttt{q501} to 26/30 in \texttt{q601}. Qwen3 4B (thinking) improved from consistently below 20 to an average of 28.375. Most of the other models got perfect scores with excellent consistency.

This illustrates the best case result of using failure mode analysis to improve an agentic AI deployment. The cost of LLM inference needed for the task can plummet, while simultaneously increasing overall accuracy and reliability.

\subsubsection{But Sometimes Hints Can Lower Performance!}

Not all hints work the same for all models or scenarios. We applied the same methodology to create \texttt{q602}, starting with failure mode analysis for \texttt{q502}. \texttt{q502} is similar to \texttt{q501}, except this time 6 facts are requested from the LLM agent to be retrieved from the database, instead of just one. 

We applied the same modifications from \texttt{q601}, plus another instruction that explicitly informs the LLM that a zero value can be a legitimate answer for one of the facts requested if the randomized company data is truly not in the dataset, to mitigate an LLM behavior where encountering a 0 query result causes them to think the query is wrong and sometimes leads them to decide on a different query filter - which of course leads to a failed test.

Due to the difficulty of this question, the existing average for \texttt{q502} was low: 5.84 out of 30 across all models. This was only modestly improved to 7.33, but the effect was not consistent across all models. For example, our overall best model, Qwen3 235B-A22B Instruct 2507, suffered a performance decrease, shown in \ref{tab:hints-impact}.

Failure mode analysis revealed that the hints caused two distinct negative effects on model behavior:

\textbf{Effect 1: Increased Capitalization Errors.} The hints triggered Qwen3 235B-A22B Instruct 2507 to shift from copying exact values from the instruction text (which were lowercase, like ``active'') to applying assumed database naming conventions (capitalizing values, as in ``Active''). This ``database convention mode'' caused capitalization errors to increase dramatically by 13-16\%.

\textbf{Effect 2: Suppression of the self-correction mechanism.} Having the hints in place completely eliminated a self-correction behavior where Qwen3 235B-A22B would sample actual database values to verify their assumptions:
\begin{itemize}
    \item Q502: 15/480 samples (3.1\%) attempted data sampling with 93.3\% success rate
    \item Q602: 0/480 samples (0\%)---mechanism completely suppressed
\end{itemize}

These results demonstrate that even well-intentioned hints can trigger unintended mental model shifts in LLMs. The schema hint caused models to seemingly prioritize perceived database conventions over literal instruction text, while simultaneously eliminating any tendency to verify assumptions through data sampling. This highlights the need for careful empirical validation of prompting strategies, as interventions that seem benign and designed to be helpful may actually degrade performance through subtle behavioral changes.

\begin{table}[ht]
\centering
\caption{Impact of Hints on Model Performance: Q502 vs Q602}
\label{tab:hints-impact}
\begin{tabular}{lcccc}
\toprule
& \multicolumn{2}{c}{\textbf{Q502 (Standard)}} & \multicolumn{2}{c}{\textbf{Q602 (Easy/Guided)}} \\
\cmidrule(lr){2-3} \cmidrule(lr){4-5}
\textbf{Model} & \textbf{Success Rate} & \textbf{Correct/Total} & \textbf{Success Rate} & \textbf{Correct/Total} \\
\midrule
qwen3\_235b\_a22b\_instruct\_2507\_fp8\_014813 & 75.4\% & 181/240 & 71.7\% & 172/240 \\
qwen3\_235b\_a22b\_instruct\_2507\_003419 & 71.7\% & 172/240 & 56.2\% & 135/240 \\
qwen3\_235b\_a22b\_instruct\_2507\_083805 & 67.9\% & 163/240 & 58.8\% & 141/240 \\
qwen3\_235b\_a22b\_003259 & 45.8\% & 110/240 & 55.0\% & 132/240 \\
\bottomrule
\end{tabular}
\end{table}

\subsubsection{Original Qwen3 ``Hybrid Thinking'' Models Severely Underperformed Expectations}

The original Qwen3 launched at the end of April 2025), showing impressive benchmarks \cite{qwen2025qwen3} \cite{yang2025qwen3technicalreport}.

As previously noted in \ref{sec:benchmark-comparison}, we found that the Qwen3 models originally released, featuring hybrid thinking, actually fell flat when compared against Qwen2.5.

Table \ref{tab:qwen-v-qwen} shows a filtered view of KAMI v0.1 showing only Qwen2.5 and Qwen3 models (pre-2507 refresh).

Even the biggest model released - Qwen3 235B-A22B (non-thinking-mode) - fared worse in KAMI v0.1 than Qwen2.5 72B Instruct. In fact, even when in thinking mode, none of the dense models (from 4B to 32B) were any better than Qwen2.5 72B Instruct.

Worse, most of the Qwen3 dense models underperform even Qwen2.5 14B Instruct! (It should be noted that, in what seems to be a pattern that holds true for both generations of Qwen, the 14B version is decidedly better than the 32B - another instance of ``bigger is not always better''). In non-thinking mode, only Qwen3 235B-A22B beats Qwen2.5 14B. Qwen3 4B and 8B, even in thinking mode, scores significantly lower than Qwen2.5 14B.

It is no surprise then that in the next refresh (July 2025), the Qwen Team released Qwen3 variants without hybrid thinking, and instead produced dedicated Instruct (non-reasoning) and Thinking (reasoning) models, which boosted performance. Qwen3 235B-A22B Instruct 2507, for example, became the overall KAMI v0.1 leader, while its original variant barely beats Qwen2.5 14B Instruct (some t-CI overlap) and loses clearly against Qwen2.5 72B Instruct.

\begin{table}[ht]
\centering
\caption{Qwen3 vs Qwen2.5}
\label{tab:qwen-v-qwen}
\begin{threeparttable}
\footnotesize
\begin{tabular}{lp{4cm}rrrrrrrr}
\toprule
\textbf{Platform} & \textbf{Model} & \textbf{Runs} & \textbf{Pooled Acc.} & \textbf{Std Dev} & \textbf{RSE} & \textbf{95\% t-CI} & \textbf{Range} \\
\midrule
AMD MI300X & qwen3\_30b\_a3b\_thinkmode & 8 & 72.7\% & ±1.40\% & ±26.7\% & (71.6\%, 73.9\%) & 3.86\% \\
AMD MI300X & qwen2\_5\_72b\_instruct & 7 & 71.1\% & ±0.78\% & ±28.9\% & (70.4\%, 71.8\%) & 2.46\% \\
AMD MI300X & qwen3\_14b\_thinkmode & 8 & 69.1\% & ±1.54\% & ±26.7\% & (67.8\%, 70.4\%) & 4.39\% \\
Intel Gaudi3 & qwen3\_235b\_a22b & 8 & 67.7\% & ±1.17\% & ±26.7\% & (66.7\%, 68.6\%) & 3.33\% \\
AMD MI300X & qwen3\_32b\_thinkmode & 8 & 67.6\% & ±2.13\% & ±26.7\% & (65.8\%, 69.4\%) & 6.32\% \\
AMD MI300X & qwen2\_5\_14b\_instruct & 8 & 66.6\% & ±1.19\% & ±26.7\% & (65.6\%, 67.6\%) & 3.16\% \\
AMD MI300X & qwen3\_32b\_fp8 & 8 & 63.7\% & ±0.86\% & ±26.7\% & (63.0\%, 64.4\%) & 2.63\% \\
AMD MI300X & qwen3\_8b\_thinkmode & 8 & 62.5\% & ±1.81\% & ±26.7\% & (61.0\%, 64.1\%) & 5.09\% \\
AMD MI300X & qwen3\_32b & 8 & 61.6\% & ±0.89\% & ±26.7\% & (60.8\%, 62.3\%) & 2.81\% \\
AMD MI300X & qwen3\_14b\_fp8 & 8 & 60.0\% & ±0.66\% & ±26.7\% & (59.5\%, 60.5\%) & 1.75\% \\
Intel Gaudi3 & qwen3\_32b & 8 & 59.7\% & ±1.32\% & ±26.7\% & (58.6\%, 60.8\%) & 3.33\% \\
AMD MI300X & qwen3\_14b & 8 & 58.7\% & ±0.68\% & ±26.7\% & (58.2\%, 59.3\%) & 1.75\% \\
AMD MI300X & qwen3\_30b\_a3b & 8 & 58.1\% & ±0.88\% & ±26.7\% & (57.4\%, 58.8\%) & 2.28\% \\
AMD MI300X & qwen2\_5\_32b\_instruct & 8 & 55.9\% & ±1.29\% & ±26.7\% & (54.8\%, 57.0\%) & 3.86\% \\
AMD MI300X & qwen3\_4b\_thinkmode & 8 & 50.5\% & ±1.55\% & ±26.7\% & (49.2\%, 51.8\%) & 4.21\% \\
AMD MI300X & qwen3\_8b & 5 & 49.1\% & ±0.99\% & ±35.4\% & (47.8\%, 50.3\%) & 2.63\% \\
AMD MI300X & qwen2\_5\_7b\_instruct & 8 & 41.6\% & ±1.21\% & ±26.7\% & (40.5\%, 42.6\%) & 3.68\% \\
AMD MI300X & qwen3\_4b & 3 & 37.8\% & ±1.41\% & ±50.0\% & (34.3\%, 41.3\%) & 2.81\% \\
\bottomrule
\end{tabular}
\begin{tablenotes}
\footnotesize
\item Models sorted by mean accuracy. RSE (Relative Standard Error) reflects uncertainty in std dev estimates. t-CI is the 95\% confidence interval for mean accuracy using t-distribution (appropriate for small sample sizes). (see Section~\ref{sec:kami-metrics-collected}).
\end{tablenotes}
\end{threeparttable}
\end{table}

\subsubsection{FP8 vs Full-Precision}

Another interesting finding was that FP8 quantization does not seem to harm performance for Qwen3 models of different sizes, ranging from 14B to 235B. Table \ref{tab:fp8-vs-fullprec} shows the results with only the model configurations with accompanying FP8 variants shown.

\begin{table}[ht]
\centering
\caption{Full-Precision vs FP8-Quantized Variants}
\label{tab:fp8-vs-fullprec}
\begin{threeparttable}
\footnotesize
\begin{tabular}{lp{4cm}rrrrrrrr}
\toprule
\textbf{Platform} & \textbf{Model} & \textbf{Runs} & \textbf{Pooled Acc.} & \textbf{Std Dev} & \textbf{RSE} & \textbf{95\% t-CI} & \textbf{Range} \\
\midrule
AMD MI300X & qwen3\_235b\_a22b\_instruct\_2507\_fp8 & 8 & 88.8\% & ±1.19\% & ±26.7\% & (87.8\%, 89.7\%) & 3.33\% \\
Intel Gaudi3 & qwen3\_235b\_a22b\_instruct\_2507 & 8 & 88.4\% & ±1.43\% & ±26.7\% & (87.2\%, 89.6\%) & 3.51\% \\
AMD MI300X & qwen3\_235b\_a22b\_instruct\_2507 & 8 & 88.2\% & ±1.54\% & ±26.7\% & (87.0\%, 89.5\%) & 4.21\% \\
AMD MI300X & llama\_4\_maverick\_17b\_128e\_instruct & 8 & 74.6\% & ±0.93\% & ±26.7\% & (73.8\%, 75.3\%) & 2.81\% \\
AMD MI300X & llama\_3\_3\_70b\_instruct\_fp8\_kv & 8 & 74.5\% & ±1.61\% & ±26.7\% & (73.2\%, 75.9\%) & 5.26\% \\
AMD MI300X & llama\_4\_maverick\_17b\_128e\_instruct\_fp8 & 8 & 73.1\% & ±1.60\% & ±26.7\% & (71.8\%, 74.5\%) & 4.74\% \\
AMD MI300X & llama\_3\_3\_70b\_instruct & 8 & 71.6\% & ±1.51\% & ±26.7\% & (70.4\%, 72.9\%) & 5.44\% \\
AMD MI300X & qwen3\_32b\_fp8 & 8 & 63.7\% & ±0.86\% & ±26.7\% & (63.0\%, 64.4\%) & 2.63\% \\
AMD MI300X & qwen3\_32b & 8 & 61.6\% & ±0.89\% & ±26.7\% & (60.8\%, 62.3\%) & 2.81\% \\
AMD MI300X & qwen3\_14b\_fp8 & 8 & 60.0\% & ±0.66\% & ±26.7\% & (59.5\%, 60.5\%) & 1.75\% \\
AMD MI300X & qwen3\_14b & 8 & 58.7\% & ±0.68\% & ±26.7\% & (58.2\%, 59.3\%) & 1.75\% \\
\bottomrule
\end{tabular}
\begin{tablenotes}
\footnotesize
\item Main KAMI v0.1 results, filtered to show models tested with FP8-quantized variants.
\end{tablenotes}
\end{threeparttable}
\end{table}

Comparing the 95\% t-CIs, the small Qwen3 models (14B and 32B) show a slight but significant improvement (no overalp in t-CIs) with FP8 quantization. (We used the official FP8 quantized model available from the Qwen Team's Hugging Face repo)

Qwen3 235B-A22B Instruct 2507 is a slightly different story - while the pooled accuracy slightly of the FP8 variant beats either of the full-precision runs (Gaudi 3 and MI300X), there is significant overlap in t-CIs. The results do paint an ever-so-slightly better agentic performance, taking into account the tighter Standard Dev, smaller Range (difference between Min and Max scores across runs), and characteristics of the t-CI overlap (FP8 t-CI starts a little higher, and ends a little higher). In practice, such small differences would likely be undetectable.

Llama 3.3 70B Instruct shows another case where t-CIs do not overlap in favor of the FP8 variant. (We used the ``amd/Llama-3.3-70B-Instruct-FP8-KV'' model, as there was no official one from Meta). 

Llama 4 Maverick is the only model whose FP8 variant underperformed the full-precision model. (We used the official FP8 model from Meta)

The insight here is that FP8 does not always reduce or increase agentic performance. The results seem to be model-specific effects combined with use case specific factors. Not all quantizations are also necessarily the same, so FP8 performance depending on how the quantization was made may vary, hence noting which FP8 models we specifically used.

We plan to include more quantized models in future KAMI versions, given the interesting results in v0.1.

\subsubsection{Tools are Prompts and Critical to Context Engineering}
\label{sec:tools-are-prompts}
One of the failure modes we observed stemmed from the messaging of the Python execution tool for our LLM agents. This tool allows the LLM to execute arbitrary Python code. The results of running that code are then returned back to the LLM (e.g., \texttt{print} statements or debugging lines, or any error messages). In cases where the Python code produces no output (no error happened, and nothing is ever printed to the console), then the tool returns the message ``Code executed successfully with no output''.

During the KAMI v0.1 trials, we observed even the better LLMs currently assessed, such as Qwen3 235B-A22B Instruct 2507, can get confused by this message. For example, in the Q303 question template (find two words at random positions from a text file), models would often correctly identify the need to use Python. Then, despite the assurance that the ``code executed successfully'', the part that said ``no output'' caused models to think something was wrong and try again, sometimes adding \texttt{print()} statements. In some cases, the model would end up thinking there was something wrong with the Python tool altogether, and revert to just trying to manually count the words in the text file (which, due to the high positional numbers involved, LLMs never succeed at). 

The lesson here is that everything about a tool - from its description, usage information, parameters, parameter description, and even the messages it sends back during success and failure cases - is a critical part context engineering. The timely appearance of helpful or confusing messages can end up helping or hindering the performance of LLM agents in unexpected ways.

This observation aligns with Anthropic's systematic findings on tool design for LLM agents~\cite{anthropic2025tools}. Their evaluation-driven research demonstrates that tool response messaging significantly impacts agent performance. They emphasize that when tool calls raise errors or produce unexpected outputs, response messages should ``clearly communicate specific and actionable improvements, rather than opaque error codes or tracebacks''. Our experience with the ``no output'' message exemplifies this principle: what seemed like a helpful success confirmation actually introduced ambiguity that degraded performance. Anthropic's research further reveals that even small refinements to tool descriptions and messaging can yield dramatic improvements---they achieved state-of-the-art performance on SWE-bench Verified specifically through ``precise refinements to tool descriptions, dramatically reducing error rates and improving task completion''.

This insight has critical implications for enterprise deployment: off-the-shelf MCP~\cite{mcp2025} servers may suffice for prototyping and testing, but are likely to underwhelm in production. While MCP servers have gained widespread adoption, with vendors eagerly embracing the standard, our findings - supported by aligned guidance from Anthropic, who designed and popularized MCP - indicate that maximizing agentic performance will require customized toolsets. If LLM tools are ultimately prompts, then having full control of prompting to maximize use case performance means having control of the prompts contained within the tools: descriptions, parameters, syntax, messages, and even the name of the tool itself. This necessitates both control and customization capability. For enterprise agentic AI deployments that require 99.9\% or higher success rates, having full control of all context engineering opportunities is critical.

Similarly, naively wrapping APIs in MCP servers with no regard to context engineering is bound to underwhelm in production. Without full control of all context engineering possibilities within a tool - such as providing more helpful guidance during routine results or edge cases - enterprises sacrifice the performance optimizations that distinguish production-ready systems from demos and prototypes.

\subsection{Time, Latency, and Token Usage: Reasoning vs Instruct Models}
\label{sec:reasoning-vs-instruct}

From our current batch of results, deploying reasoning models for the typical agentic tasks we tested does not present itself as a good tradeoff. 

We've discussed \texttt{qwen3\_4b} in an earlier section. To recap, with reasoning disabled, it performs abysmally with a pooled accuracy of 37.8\%, with an average of 900 tokens per conversation. Enabling reasoning boosts this to 50.5\%, but at the cost of 14x more tokens. Wall-time increased by almost 6x. (See Table \ref{tab:reasoning-impact} in Section \ref{sec:results} for the timing and tokens data)

\texttt{qwen3\_8b} fares a little better. With reasoning disabled, it gets  49.1\% (comparable to \texttt{qwen3\_4b} with reasoning), with only 630 tokens per conversation on average. Enabling reasoning boosts accuracy to 62.5\%, but increases average tokens per conversation by over 10x. Wall-time increased by about 4x.

\texttt{qwen3\_14b} increases performance nicely from 58.7\% to 69.1\%, while tokens generated rose by 11x, from 448.7 on average to 4929.5. Wall-time increased by almost 6x.

\texttt{qwen3\_32b} improved from 59.7\% to 67.6\%, with a similar 10x increase in generated tokens, and a wall-time increase of over 5x. Notably, while its non-reasoning score was only barely better than \texttt{qwen3\_14b} its reasoning score was slightly worse. However, in both cases, there is significant overlap in their 95\% t-CI. Disappointingly, this means it would barely be useful to run qwen3\_32b at all for our set of agentic tasks, since the smaller 14B variant is always same or better.

\texttt{qwen3\_30b\_a3b\_thinkmode} gets the biggest boost, from 58.1\% to 72.7\%, with over 11x more tokens and a wall-time increase of over 6x.

Overall, reasoning models we tested increase task latency by 4-6x, and increase output tokens by 10-14x, summarized in Table~\ref{tab:reasoning-impact}. This is a steep price to pay, especially for per-token usage. It is hard to justify based on the numbers we see in KAMI v0.1. Finding an applicable non-reasoning model will be better in terms of latency and token cost. For self-hosted models where per-token pricing does not apply, saving 10x output tokens still means less overall GPU utilization, which translates to higher total effective GPU capacity available.

\begin{table}[ht]
\centering
\caption{Impact of Reasoning Mode on Model Performance}
\label{tab:reasoning-impact}
\begin{tabular}{lccc}
\hline
\textbf{Model} & \textbf{Accuracy Gain} & \textbf{Token Increase} & \textbf{Wall-time Increase} \\
\hline
Qwen3 4B & 37.8\% $\rightarrow$ 50.5\% & 14$\times$ & 6$\times$ \\
Qwen3 8B & 49.1\% $\rightarrow$ 62.5\% & 10$\times$ & 4$\times$ \\
Qwen3 14B & 58.7\% $\rightarrow$ 69.1\% & 11$\times$ & 6$\times$ \\
Qwen3 32B & 59.7\% $\rightarrow$ 67.6\% & 10$\times$ & 5$\times$ \\
Qwen3 30B-A3B & 58.1\% $\rightarrow$ 72.7\% & 11$\times$ & 6$\times$ \\
\hline
\end{tabular}
\end{table}

\section{Conclusion}
\label{sec:conclusion}

The Kamiwaza Agentic Merit Index (KAMI) v0.1 demonstrates that traditional LLM benchmarks often fail to predict real-world agentic performance in common, mundane enterprise settings. Through the evaluation of 35 model configurations across over 5.5 billion tokens and 170,000 test items, we reveal a persistent ``agentic disconnect'': models that excel on standard or even aggregated benchmarks may underperform on mundane but critical enterprise tasks like CSV analysis, database querying, and file manipulation. Conversely, some models dismissed by conventional metrics such as Llama 3.1 70B or Claude 3.5 Haiku show strong, reliable performance in practical agentic workflows.

Our results challenge several assumptions common in the field: that newer models are immediately universally better (Qwen3 models underperforming against Qwen2.5 72B and 14B), that larger parameter counts guarantee superior agentic capability (ef., Qwen2.5 and Qwen3 32B models underperform their respective 14B variants), and that reasoning-augmented architectures always justify their computational overhead. Instead, KAMI v0.1 highlights the importance of task-specific evaluation, reliability metrics, and contamination-resistant design—principles embodied in the underlying PICARD framework. KAMI v0.1 results also highlight that small interventions in prompt design, tool feedback, and environment structuring can have disproportionately large effects on both performance and cost - underscoring that context engineering is as important as model selection.

We view KAMI not as an endpoint but as the foundation of a long-term effort. Looking ahead, KAMI aims to become the ``SPEC CPU for agentic AI'': a trusted, standardized benchmark that reflects the repetitive, scoped, multi-step, and tool-dependent nature of real enterprise agentic AI deployments. Future versions will expand into vendor-specific environments (e.g., Oracle, MongoDB), increase model coverage (e.g., Llama-3.1-405B, DeepSeek, GLM-4.6, proprietary models) incorporate more granular reliability metrics, and explore the impact of custom tool design on agent performance. By grounding evaluation in practical utility rather than leaderboard optics, KAMI offers enterprises a clearer path to successful, cost-effective agentic AI adoption.

Data Availability: The raw result files will be made available at \href{https://docs.kamiwaza.ai/research/datasets}{https://docs.kamiwaza.ai/research/datasets}.

\section*{Acknowledgements}

This research was made possible through the generous provision of compute resources by Signal65, who provided access to 32 AMD MI300X GPUs across four servers for experimental evaluation. We thank Ryan Shrout, Brian Martin, Mitch Lewis, and Russ Fellows for their support in enabling this large-scale benchmarking effort.

\section*{AI Usage Disclosure}

The researchers used the following generative AI services to assist with the manuscript:

\begin{itemize}
    \item Qwen Chat: Qwen3 235B-A22B, Qwen3 Max
    \item Claude Desktop: Claude Sonnet 4.5
    \item ChatGPT: GPT 5 Instant
\end{itemize}

About 70\% of the content was written directly by human researchers. As each section was written, we sometimes ask generative AI for review and suggestions. We routinely accept only about 50\% of the suggestions. The suggestions we don't accept are either very low impact or stylistic changes that we do not prefer, or simply hallucinations or misunderstandings of the text.

About 30\% of the content was generated directly through generative AI. Both the Abstract and Conclusion were generated from generative AI by feeding it the existing manuscript. In addition, all tables were created through generative AI directly using raw source data. For all of these sections and elements generated directly by generative AI, we use one of Claude Desktop, Qwen Chat, and ChatGPT to generate the first draft, and then have the other two do reviews. Then, human researchers do a final review, editing and integration. For the Conclusion in particular, we had three models (Claude Sonnet 4.5, Qwen3 Max, GPT 5 Instant)  generate their own Conclusion section based on the full manuscript. Human researchers then did the review, editing and integration.

In all cases, final editorial control, technical validation, and intellectual responsibility rest solely with the human authors. The authors take full responsibility for the accuracy and integrity of all content in this manuscript.

\bibliography{references}

\clearpage
\appendix
\renewcommand{\thesection}{Appendix \Alph{section}}

\section{Complete KAMI v0.1 Test Suite}
\label{app:test-suite}

The full KAMI v0.1 test definitions of all 19 question templates are shown below.

\begin{lstlisting}[language=yaml, caption={Kamiwaza Agentic Merit Index (KAMI) v0.1}, label={lst:KAMIv0.1}]
tests:

  #SECTION 100: Sanity check - Do models end up using tools unnecessarily?

  - question_id: 101
    samples: 30
    template: "Respond only with this word: {{entity1}}"
    scoring_type: "stringmatch"
    expected_response: "{{entity1}}"

  - question_id: 102
    samples: 30
    template: "Respond only with these three words in the given order: {{entity1}} {{entity2}} {{entity3}}"
    scoring_type: "stringmatch"
    expected_response: "{{entity1}} {{entity2}} {{entity3}}"

  #SECTION 200: File System

  - question_id: 201
    samples: 30
    template: "Create the following blank files: {{entity1}}.log and {{entity2}}.config in the {{artifacts}}/{{qs_id}}/{{entity3}} directory."
    scoring_type: "files_exist"
    files_to_check:
      - "{{artifacts}}/{{qs_id}}/{{entity3}}/{{entity1}}.log"
      - "{{artifacts}}/{{qs_id}}/{{entity3}}/{{entity2}}.config"

  - question_id: 202
    samples: 30
    template: |
      Create this directory structure, including all blank files specified, inside the folder `{{artifacts}}/{{qs_id}}`: 
      {{expected_structure}}
    scoring_type: "directory_structure"
    expected_structure:
      - "{{artifacts}}/{{qs_id}}/{{entity1}}/"
      - "{{artifacts}}/{{qs_id}}/{{entity1}}/{{entity2}}/"
      - "{{artifacts}}/{{qs_id}}/{{entity1}}/logs/"
      - "{{artifacts}}/{{qs_id}}/{{entity1}}/logs/{{entity3}}.log"
      - "{{artifacts}}/{{qs_id}}/{{entity4}}/"
      - "{{artifacts}}/{{qs_id}}/{{entity4}}/README.md"
      - "{{artifacts}}/{{qs_id}}/{{entity5}}/"
      - "{{artifacts}}/{{qs_id}}/{{entity5}}/{{entity6}}.{{entity7}}"
      - "{{artifacts}}/{{qs_id}}/{{entity5}}/{{entity6}}.{{entity8}}"
      - "{{artifacts}}/{{qs_id}}/{{entity5}}/{{entity6}}.{{entity9}}"

  ######################################
  #SECTION 300: Finding needles in files
  - question_id: 301
    samples: 30
    template: |
      Retrieve the following lines from this large text file {{artifacts}}/{{qs_id}}/{{entity1}}/{{entity2}}.txt:
      - "line_a": Line {{number1:1:40}}
      - "line_b": Line {{number2:51:90}}

      Write your response in a json file {{artifacts}}/{{qs_id}}/lines.json, using "line_a" and "line_b" as the keys.
      Each value should contain only the text content of that line, with no additional formatting.
    scoring_type: "readfile_jsonmatch"
    file_to_read: "{{artifacts}}/{{qs_id}}/lines.json"
    expected_content: |
      {
        "line_a": "{{file_line:{{number1:1:40}}:TARGET_FILE[notes_file]}}",
        "line_b": "{{file_line:{{number2:51:90}}:TARGET_FILE[notes_file]}}"
      }
    sandbox_setup:
      components:
        - type: "create_files"
          name: "notes_file"
          target_file: "{{artifacts}}/{{qs_id}}/{{entity1}}/{{entity2}}.txt"
          content:
            type: "lorem_lines"
            count: 100

  - question_id: 302
    samples: 30
    template: |
      Retrieve the following lines from this large text file {{artifacts}}/{{qs_id}}/{{entity1}}/{{entity2}}.txt:
      - "line_a": Line {{number1:1:15}}
      - "line_b": Line {{number2:16:30}}
      - "line_c": Line {{number3:31:45}}
      - "line_d": Line {{number4:46:60}}
      - "line_e": Line {{number5:61:75}}
      - "line_f": Line {{number6:76:90}}
      - "line_g": Line {{number7:91:105}}

      Write your response in a json file {{artifacts}}/{{qs_id}}/lines.json, using "line_a", "line_b" ... until "line_g" as the keys.
      Each value should contain only the text content of that line, with no additional formatting.
    scoring_type: "readfile_jsonmatch"
    file_to_read: "{{artifacts}}/{{qs_id}}/lines.json"
    expected_content: |
      {
        "line_a": "{{file_line:{{number1:1:15}}:TARGET_FILE[notes_file]}}",
        "line_b": "{{file_line:{{number2:16:30}}:TARGET_FILE[notes_file]}}",
        "line_c": "{{file_line:{{number3:31:45}}:TARGET_FILE[notes_file]}}",
        "line_d": "{{file_line:{{number4:46:60}}:TARGET_FILE[notes_file]}}",
        "line_e": "{{file_line:{{number5:61:75}}:TARGET_FILE[notes_file]}}",
        "line_f": "{{file_line:{{number6:76:90}}:TARGET_FILE[notes_file]}}",
        "line_g": "{{file_line:{{number7:91:105}}:TARGET_FILE[notes_file]}}"
      }
    sandbox_setup:
      components:
        - type: "create_files"
          name: "notes_file"
          target_file: "{{artifacts}}/{{qs_id}}/{{entity1}}/{{entity2}}.txt"
          content:
            type: "lorem_lines"
            count: 110


  - question_id: 303
    samples: 30
    template: |
      Retrieve the following words from this large text file {{artifacts}}/{{qs_id}}/{{entity1}}/{{entity2}}.txt:
      - "word_a": Word #{{number1:1:100}}
      - "word_b": Word #{{number2:101:200}}

      Write your response in a json file {{artifacts}}/{{qs_id}}/words.json, using "word_a" and "word_b" as the keys.
      Each value should contain only the word itself, with no additional formatting.
    scoring_type: "readfile_jsonmatch"
    file_to_read: "{{artifacts}}/{{qs_id}}/words.json"
    expected_content: |
      {
        "word_a": "{{file_word:{{number1:1:100}}:TARGET_FILE[notes_file]}}",
        "word_b": "{{file_word:{{number2:101:200}}:TARGET_FILE[notes_file]}}"
      }
    sandbox_setup:
      components:
        - type: "create_files"
          name: "notes_file"
          target_file: "{{artifacts}}/{{qs_id}}/{{entity1}}/{{entity2}}.txt"
          content:
            type: "lorem_lines"
            count: 30


  - question_id: 304
    samples: 30
    template: |
      Retrieve the following words from this large text file {{artifacts}}/{{qs_id}}/{{entity1}}/{{entity2}}.txt:
      - "word_a": Word #{{number1:1:100}}
      - "word_b": Word #{{number2:101:200}}
      - "word_c": Word #{{number3:201:300}}
      - "word_d": Word #{{number4:301:400}}
      - "word_e": Word #{{number5:401:500}}
      - "word_f": Word #{{number6:501:600}}
      - "word_g": Word #{{number7:601:700}}

      Write your response in a json file {{artifacts}}/{{qs_id}}/words.json, using "word_a", "word_b" ... until "word_g" as the keys.
      Each value should contain only the word itself, with no additional formatting.
    scoring_type: "readfile_jsonmatch"
    file_to_read: "{{artifacts}}/{{qs_id}}/words.json"
    expected_content: |
      {
        "word_a": "{{file_word:{{number1:1:100}}:TARGET_FILE[notes_file]}}",
        "word_b": "{{file_word:{{number2:101:200}}:TARGET_FILE[notes_file]}}",
        "word_c": "{{file_word:{{number3:201:300}}:TARGET_FILE[notes_file]}}",
        "word_d": "{{file_word:{{number4:301:400}}:TARGET_FILE[notes_file]}}",
        "word_e": "{{file_word:{{number5:401:500}}:TARGET_FILE[notes_file]}}",
        "word_f": "{{file_word:{{number6:501:600}}:TARGET_FILE[notes_file]}}",
        "word_g": "{{file_word:{{number7:601:700}}:TARGET_FILE[notes_file]}}"
      }
    sandbox_setup:
      components:
        - type: "create_files"
          name: "notes_file"
          target_file: "{{artifacts}}/{{qs_id}}/{{entity1}}/{{entity2}}.txt"
          content:
            type: "lorem_lines"
            count: 90


  ############################
  #Section 400: CSV Processing
  #############################
  - question_id: 401
    samples: 30
    template: "Create a JSON summary of {{artifacts}}/{{qs_id}}/{{entity1}}/crm_export.csv showing total customers (integer) and average age (full floating-point precision, do not round). Create {{artifacts}}/{{qs_id}}/cust.json with the JSON containing `total_customers` and `average_age`."
    scoring_type: "readfile_jsonmatch"
    file_to_read: "{{artifacts}}/{{qs_id}}/cust.json"
    expected_content: '{"total_customers": {{csv_count:C_ID:TARGET_FILE[crm_data]}}, "average_age": {{csv_avg:AGE_YRS:TARGET_FILE[crm_data]}}}'
    sandbox_setup:
      components:
        - type: "create_csv"
          name: "crm_data"
          target_file: "{{artifacts}}/{{qs_id}}/{{entity1}}/crm_export.csv"
          content:
            headers: ["C_ID", "C_NAME", "AGE_YRS", "LOC_CD", "REG_DT"]
            header_types: ["id", "person_name", "age", "city", "date"]
            rows: 75

  - question_id: 402
    samples: 30
    template: "Analyze the business data across multiple CSV files in {{artifacts}}/{{qs_id}}/ and create a comprehensive report at {{artifacts}}/{{qs_id}}/multi_source_analysis.json. Answer these questions: (1) How many contacts do we have in the {{semantic1:category}} industry in contacts.csv? (2) What is the average base price for {{semantic2:category}} category products in products.csv (full floating-point precision, do not round)? (3) How many customers do we have in the {{semantic3:region}} region? (4) What is the total order value for orders above {{number1:15000:35000:currency}} in orders.csv? (Include decimal point, e.g., 718313.0). (5) How many {{semantic4:status}} status orders are there in orders.csv? (6) What is the average quantity across all orders in orders.csv (full floating-point precision, do not round)? Create JSON with keys: 'industry_contact_count', 'category_avg_price', 'region_customer_count', 'high_value_total', 'status_order_count', 'avg_order_quantity'."
    scoring_type: "readfile_jsonmatch"
    file_to_read: "{{artifacts}}/{{qs_id}}/multi_source_analysis.json"
    expected_content: "{\"industry_contact_count\": {{csv_count_where:COMP_ID:INDUSTRY:==:{{semantic1:category}}:TARGET_FILE[companies_csv]}}, \"category_avg_price\": {{csv_avg_where:BASE_PRICE:CATEGORY:==:{{semantic2:category}}:TARGET_FILE[products_csv]}}, \"region_customer_count\": {{csv_count_where:CUSTOMER_ID:REGION:==:{{semantic3:region}}:TARGET_FILE[customers_csv]}}, \"high_value_total\": {{csv_sum_where:ORDER_AMOUNT:ORDER_AMOUNT:>:{{number1:15000:35000:currency}}:TARGET_FILE[orders_csv]}}, \"status_order_count\": {{csv_count_where:ORDER_ID:STATUS:==:{{semantic4:status}}:TARGET_FILE[orders_csv]}}, \"avg_order_quantity\": {{csv_avg:QUANTITY:TARGET_FILE[orders_csv]}}}"
    sandbox_setup:
      components:
        - type: "create_csv"
          name: "companies_csv"
          target_file: "{{artifacts}}/{{qs_id}}/contacts.csv"
          content:
            headers: ["COMP_ID", "COMPANY", "INDUSTRY", "CONTACT_PERSON"]
            header_types: ["id", "company", "category", "person_name"]
            rows: "{{number2:40:50}}"
        
        - type: "create_csv"
          name: "products_csv"
          target_file: "{{artifacts}}/{{qs_id}}/products.csv"
          content:
            headers: ["PROD_ID", "PROD_NAME", "CATEGORY", "BASE_PRICE", "SUPPLIER"]
            header_types: ["id", "product", "category", "price", "company"]
            rows: "{{number3:60:70}}"
        
        - type: "create_csv"
          name: "customers_csv"
          target_file: "{{artifacts}}/{{qs_id}}/customers.csv"
          content:
            headers: ["CUSTOMER_ID", "CUSTOMER_NAME", "REGION", "DEPARTMENT", "SIGNUP_DATE"]
            header_types: ["id", "person_name", "region", "department", "date"]
            rows: "{{number4:40:50}}"
        
        - type: "create_csv"
          name: "orders_csv"
          target_file: "{{artifacts}}/{{qs_id}}/orders.csv"
          content:
            headers: ["ORDER_ID", "ORDER_AMOUNT", "QUANTITY", "STATUS", "ORDER_DATE"]
            header_types: ["id", "currency", "score", "status", "date"]
            rows: "{{number5:40:50}}"

  - question_id: 403
    samples: 30
    template: "Analyze the business data files in {{artifacts}}/{{qs_id}}/ and determine: What is the total value of all orders placed by customers from the {{semantic1:region}} region? (Include decimal point, e.g., 718313.0). Save your answer as a JSON file at {{artifacts}}/{{qs_id}}/regional_analysis.json with the key 'total_regional_value'."
    scoring_type: "readfile_jsonmatch"
    file_to_read: "{{artifacts}}/{{qs_id}}/regional_analysis.json"
    expected_content: "{\"total_regional_value\": {{csv_sum_where:ORDER_AMOUNT:CUSTOMER_REGION:==:{{semantic1:region}}:TARGET_FILE[orders_with_regions_csv]}}}"
    sandbox_setup:
      components:
        - type: "create_csv"
          name: "companies_csv"
          target_file: "{{artifacts}}/{{qs_id}}/contacts.csv"
          content:
            headers: ["COMP_ID", "COMPANY", "INDUSTRY", "CONTACT_PERSON"]
            header_types: ["id", "company", "category", "person_name"]
            rows: "{{number2:40:50}}"
        
        - type: "create_csv"
          name: "products_csv"
          target_file: "{{artifacts}}/{{qs_id}}/products.csv"
          content:
            headers: ["PROD_ID", "PROD_NAME", "CATEGORY", "BASE_PRICE", "SUPPLIER"]
            header_types: ["id", "product", "category", "price", "company"]
            rows: "{{number3:40:50}}"
        
        - type: "create_csv"
          name: "customers_csv"
          target_file: "{{artifacts}}/{{qs_id}}/customers.csv"
          content:
            headers: ["CUSTOMER_ID", "CUSTOMER_NAME", "REGION", "DEPARTMENT", "SIGNUP_DATE"]
            header_types: ["id", "person_name", "region", "department", "date"]
            rows: "{{number4:40:50}}"
        
        - type: "create_csv"
          name: "orders_with_regions_csv"
          target_file: "{{artifacts}}/{{qs_id}}/orders.csv"
          content:
            headers: ["ORDER_ID", "ORDER_AMOUNT", "QUANTITY", "STATUS", "CUSTOMER_REGION"]
            header_types: ["id", "currency", "score", "status", "region"]
            rows: "{{number5:100:150}}"


  #################################
  #Section 500: Database processing
  - question_id: 501
    samples: 30
    template: "How many orders above {{number1:10000:20000:currency}} are there from customers in the {{semantic2:region}} region in {{artifacts}}/{{qs_id}}/{{entity1}}.db? Create a JSON file {{artifacts}}/{{qs_id}}/big_orders_count.json that contains the answer using 'num_big_orders' as key."
    scoring_type: "readfile_jsonmatch"
    file_to_read: "{{artifacts}}/{{qs_id}}/big_orders_count.json"
    expected_content: "{\"num_big_orders\": {{sqlite_query:SELECT COUNT(*) FROM enterprise_orders o JOIN enterprise_customers c ON o.CUST_REF = c.CUST_ID WHERE c.LOC_CD = '{{semantic2:region}}' AND o.ORD_AMT > {{number1:10000:20000:currency}}:TARGET_FILE[order_db]}}}"
    sandbox_setup:
      components:
        - type: "create_sqlite"
          name: "order_db"
          target_file: "{{artifacts}}/{{qs_id}}/{{entity1}}.db"
          content:
            tables:
              - name: "enterprise_companies"
                columns:
                  - {name: "COMP_ID", type: "auto_id"}
                  - {name: "COMP_NM", type: "TEXT", data_type: "company"}
                  - {name: "INDUSTRY", type: "TEXT", data_type: "category"}
                rows: "{{number2:8:15}}"

              - name: "enterprise_products"
                columns:
                  - {name: "PROD_ID", type: "auto_id"}
                  - {name: "PROD_NM", type: "TEXT", data_type: "product"}
                  - {name: "CATEGORY", type: "TEXT", data_type: "category"}
                  - {name: "BASE_PRICE", type: "INTEGER", data_type: "price"}
                rows: "{{number3:20:30}}"

              - name: "enterprise_customers"
                columns:
                  - {name: "CUST_ID", type: "auto_id"}
                  - {name: "CUST_NM", type: "TEXT", data_type: "person_name"}
                  - {name: "COMP_REF", type: "INTEGER", foreign_key: "enterprise_companies.COMP_ID"}
                  - {name: "LOC_CD", type: "TEXT", data_type: "region"}
                  - {name: "DEPT_CD", type: "TEXT", data_type: "department"}
                rows: "{{number4:200:250}}"

              - name: "enterprise_orders"
                columns:
                  - {name: "ORD_ID", type: "auto_id"}
                  - {name: "CUST_REF", type: "INTEGER", foreign_key: "enterprise_customers.CUST_ID"}
                  - {name: "PROD_REF", type: "INTEGER", foreign_key: "enterprise_products.PROD_ID"}
                  - {name: "ORD_AMT", type: "INTEGER", data_type: "currency"}
                  - {name: "QUANTITY", type: "INTEGER", data_type: "score"}
                  - {name: "STAT_CD", type: "TEXT", data_type: "status"}
                rows: "{{number5:750:1000}}"


  - question_id: 502
    samples: 30
    template: "Analyze the database {{artifacts}}/{{qs_id}}/{{entity1}}.db and create a comprehensive business report at {{artifacts}}/{{qs_id}}/business_analysis.json with the following information: (1) How many orders above {{number1:15000:35000:currency}} are there from {{semantic1:company}} customers in the {{semantic2:region}} region? (2) What is the total order value for all {{semantic3:department}} department customers? (3) How many unique products have been ordered by customers from {{semantic1:company}}? (4) What is the average order amount for {{semantic4:category}} category products (rounded to 2 decimal places)? (5) How many customers have made orders with quantities above {{number2:70:85}}? (6) What is the total number of {{semantic5:status}} status orders? Create a JSON with keys: 'high_value_orders', 'dept_total_value', 'company_unique_products', 'category_avg_amount', 'high_quantity_customers', and 'status_order_count'."
    scoring_type: "readfile_jsonmatch"
    file_to_read: "{{artifacts}}/{{qs_id}}/business_analysis.json"
    expected_content: "{\"high_value_orders\": {{sqlite_query:SELECT COUNT(*) FROM enterprise_orders o JOIN enterprise_customers c ON o.CUST_REF = c.CUST_ID JOIN enterprise_companies comp ON c.COMP_REF = comp.COMP_ID WHERE comp.COMP_NM = '{{semantic1:company}}' AND c.LOC_CD = '{{semantic2:region}}' AND o.ORD_AMT > {{number1:15000:35000:currency}}:TARGET_FILE[order_db]}}, \"dept_total_value\": {{sqlite_query:SELECT COALESCE(SUM(o.ORD_AMT), 0) FROM enterprise_orders o JOIN enterprise_customers c ON o.CUST_REF = c.CUST_ID WHERE c.DEPT_CD = '{{semantic3:department}}':TARGET_FILE[order_db]}}, \"company_unique_products\": {{sqlite_query:SELECT COUNT(DISTINCT o.PROD_REF) FROM enterprise_orders o JOIN enterprise_customers c ON o.CUST_REF = c.CUST_ID JOIN enterprise_companies comp ON c.COMP_REF = comp.COMP_ID WHERE comp.COMP_NM = '{{semantic1:company}}':TARGET_FILE[order_db]}}, \"category_avg_amount\": {{sqlite_query:SELECT COALESCE(ROUND(AVG(o.ORD_AMT), 2), 0) FROM enterprise_orders o JOIN enterprise_products p ON o.PROD_REF = p.PROD_ID WHERE p.CATEGORY = '{{semantic4:category}}':TARGET_FILE[order_db]}}, \"high_quantity_customers\": {{sqlite_query:SELECT COUNT(DISTINCT c.CUST_ID) FROM enterprise_orders o JOIN enterprise_customers c ON o.CUST_REF = c.CUST_ID WHERE o.QUANTITY > {{number2:70:85}}:TARGET_FILE[order_db]}}, \"status_order_count\": {{sqlite_query:SELECT COUNT(*) FROM enterprise_orders WHERE STAT_CD = '{{semantic5:status}}':TARGET_FILE[order_db]}}}"
    sandbox_setup:
      components:
        - type: "create_sqlite"
          name: "order_db"
          target_file: "{{artifacts}}/{{qs_id}}/{{entity1}}.db"
          content:
            tables:
              - name: "enterprise_companies"
                columns:
                  - {name: "COMP_ID", type: "auto_id"}
                  - {name: "COMP_NM", type: "TEXT", data_type: "company"}
                  - {name: "INDUSTRY", type: "TEXT", data_type: "category"}
                rows: "{{number3:8:15}}"

              - name: "enterprise_products"
                columns:
                  - {name: "PROD_ID", type: "auto_id"}
                  - {name: "PROD_NM", type: "TEXT", data_type: "product"}
                  - {name: "CATEGORY", type: "TEXT", data_type: "category"}
                  - {name: "BASE_PRICE", type: "INTEGER", data_type: "price"}
                rows: "{{number4:60:70}}"

              - name: "enterprise_customers"
                columns:
                  - {name: "CUST_ID", type: "auto_id"}
                  - {name: "CUST_NM", type: "TEXT", data_type: "person_name"}
                  - {name: "COMP_REF", type: "INTEGER", foreign_key: "enterprise_companies.COMP_ID"}
                  - {name: "LOC_CD", type: "TEXT", data_type: "region"}
                  - {name: "DEPT_CD", type: "TEXT", data_type: "department"}
                rows: "{{number5:200:250}}"

              - name: "enterprise_orders"
                columns:
                  - {name: "ORD_ID", type: "auto_id"}
                  - {name: "CUST_REF", type: "INTEGER", foreign_key: "enterprise_customers.CUST_ID"}
                  - {name: "PROD_REF", type: "INTEGER", foreign_key: "enterprise_products.PROD_ID"}
                  - {name: "ORD_AMT", type: "INTEGER", data_type: "currency"}
                  - {name: "QUANTITY", type: "INTEGER", data_type: "score"}
                  - {name: "STAT_CD", type: "TEXT", data_type: "status"}
                rows: "{{number6:750:1000}}"


  - question_id: 503
    samples: 30
    template: "Analyze the business database at {{artifacts}}/{{qs_id}}/{{entity1}}.db and determine: What is the total revenue generated from {{semantic1:category}} category products sold to customers in the {{semantic2:region}} region? Save your answer as a JSON file at {{artifacts}}/{{qs_id}}/category_regional_revenue.json with the key 'total_category_regional_revenue'."
    scoring_type: "readfile_jsonmatch"
    file_to_read: "{{artifacts}}/{{qs_id}}/category_regional_revenue.json"
    expected_content: "{\"total_category_regional_revenue\": {{sqlite_query:SELECT COALESCE(SUM(o.ORDER_AMT), 0) FROM orders o JOIN customers c ON o.CUSTOMER_ID = c.CUSTOMER_ID JOIN products p ON o.PRODUCT_ID = p.PRODUCT_ID WHERE p.CATEGORY = '{{semantic1:category}}' AND c.REGION = '{{semantic2:region}}':TARGET_FILE[business_db]}}}"
    sandbox_setup:
      components:
        - type: "create_sqlite"
          name: "business_db"
          target_file: "{{artifacts}}/{{qs_id}}/{{entity1}}.db"
          content:
            tables:
              # DISTRACTOR: Company info (not needed for the query)
              - name: "companies"
                columns:
                  - {name: "COMPANY_ID", type: "auto_id"}
                  - {name: "COMPANY_NAME", type: "TEXT", data_type: "company"}
                  - {name: "INDUSTRY", type: "TEXT", data_type: "category"}
                rows: "{{number2:10:20}}"
              
              # DISTRACTOR: Employee data (not needed)
              - name: "employees"
                columns:
                  - {name: "EMPLOYEE_ID", type: "auto_id"}
                  - {name: "EMPLOYEE_NAME", type: "TEXT", data_type: "person_name"}
                  - {name: "COMPANY_ID", type: "INTEGER", foreign_key: "companies.COMPANY_ID"}
                  - {name: "DEPARTMENT", type: "TEXT", data_type: "department"}
                  - {name: "SALARY", type: "INTEGER", data_type: "salary"}
                rows: "{{number3:50:200}}"
              
              # REQUIRED: Customer data (has REGION)
              - name: "customers"
                columns:
                  - {name: "CUSTOMER_ID", type: "auto_id"}
                  - {name: "CUSTOMER_NAME", type: "TEXT", data_type: "person_name"}
                  - {name: "REGION", type: "TEXT", data_type: "region"}
                  - {name: "DEPARTMENT", type: "TEXT", data_type: "department"}
                rows: "{{number4:50:100}}"
              
              # REQUIRED: Product data (has CATEGORY)
              - name: "products"
                columns:
                  - {name: "PRODUCT_ID", type: "auto_id"}
                  - {name: "PRODUCT_NAME", type: "TEXT", data_type: "product"}
                  - {name: "CATEGORY", type: "TEXT", data_type: "category"}
                  - {name: "VARIANT", type: "TEXT", data_type: "entity_pool"}
                  - {name: "MODEL", type: "TEXT", data_type: "course"}
                  - {name: "BASE_PRICE", type: "INTEGER", data_type: "price"}
                rows: "{{number5:70:100}}"
              
              # DISTRACTOR: Supplier data (not needed)
              - name: "suppliers"
                columns:
                  - {name: "SUPPLIER_ID", type: "auto_id"}
                  - {name: "SUPPLIER_NAME", type: "TEXT", data_type: "company"}
                  - {name: "CONTACT_PERSON", type: "TEXT", data_type: "person_name"}
                  - {name: "REGION", type: "TEXT", data_type: "region"}
                rows: "{{number6:50:100}}"
              
              # REQUIRED: Orders data (connects customers and products, has revenue)
              - name: "orders"
                columns:
                  - {name: "ORDER_ID", type: "auto_id"}
                  - {name: "CUSTOMER_ID", type: "INTEGER", foreign_key: "customers.CUSTOMER_ID"}
                  - {name: "PRODUCT_ID", type: "INTEGER", foreign_key: "products.PRODUCT_ID"}
                  - {name: "ORDER_AMT", type: "INTEGER", data_type: "currency"}
                  - {name: "ORDER_DATE", type: "TEXT", data_type: "date"}
                rows: "{{number7:150:200}}"


  ###############################################
  #Section 600: Database processing - Easy+Hinted

  - question_id: 601
    samples: 30
    template: "How many orders above 20000.00 are there from customers in the {{semantic2:region}} region in {{artifacts}}/{{qs_id}}/{{entity1}}.db? Create a JSON file {{artifacts}}/{{qs_id}}/big_orders_count.json that contains the answer using 'num_big_orders' as key. Begin by examining the schema to find relevant columns, and then do your analysis."
    scoring_type: "readfile_jsonmatch"
    file_to_read: "{{artifacts}}/{{qs_id}}/big_orders_count.json"
    expected_content: "{\"num_big_orders\": {{sqlite_query:SELECT COUNT(*) FROM enterprise_orders o JOIN enterprise_customers c ON o.CUST_REF = c.CUST_ID WHERE c.REGION = '{{semantic2:region}}' AND o.ORD_AMT > 20000:TARGET_FILE[order_db]}}}"
    sandbox_setup:
      components:
        - type: "create_sqlite"
          name: "order_db"
          target_file: "{{artifacts}}/{{qs_id}}/{{entity1}}.db"
          content:
            tables:
              - name: "enterprise_companies"
                columns:
                  - {name: "COMP_ID", type: "auto_id"}
                  - {name: "COMP_NM", type: "TEXT", data_type: "company"}
                  - {name: "INDUSTRY", type: "TEXT", data_type: "category"}
                rows: "{{number2:8:15}}"

              - name: "enterprise_products"
                columns:
                  - {name: "PROD_ID", type: "auto_id"}
                  - {name: "PROD_NM", type: "TEXT", data_type: "product"}
                  - {name: "CATEGORY", type: "TEXT", data_type: "category"}
                  - {name: "BASE_PRICE", type: "INTEGER", data_type: "price"}
                rows: "{{number3:20:30}}"

              - name: "enterprise_customers"
                columns:
                  - {name: "CUST_ID", type: "auto_id"}
                  - {name: "CUST_NM", type: "TEXT", data_type: "person_name"}
                  - {name: "COMP_REF", type: "INTEGER", foreign_key: "enterprise_companies.COMP_ID"}
                  - {name: "REGION", type: "TEXT", data_type: "region"}
                  - {name: "DEPT", type: "TEXT", data_type: "department"}
                rows: "{{number4:200:250}}"

              - name: "enterprise_orders"
                columns:
                  - {name: "ORD_ID", type: "auto_id"}
                  - {name: "CUST_REF", type: "INTEGER", foreign_key: "enterprise_customers.CUST_ID"}
                  - {name: "PROD_REF", type: "INTEGER", foreign_key: "enterprise_products.PROD_ID"}
                  - {name: "ORD_AMT", type: "INTEGER", data_type: "currency"}
                  - {name: "QUANTITY", type: "INTEGER", data_type: "score"}
                  - {name: "STAT_CD", type: "TEXT", data_type: "status"}
                rows: "{{number5:750:1000}}"

  - question_id: 602
    samples: 30
    template: "Analyze the database {{artifacts}}/{{qs_id}}/{{entity1}}.db and create a comprehensive business report at {{artifacts}}/{{qs_id}}/business_analysis.json with the following information: (1) How many orders above {{number1:15000:35000:currency}} are there from {{semantic1:company}} customers in the {{semantic2:region}} region? (2) What is the total order value for all {{semantic3:department}} department customers? (3) How many unique products have been ordered by customers from {{semantic1:company}}? (4) What is the average order amount for {{semantic4:category}} category products (rounded to 2 decimal places)? (5) How many customers have made orders with quantities above {{number2:70:85}}? (6) What is the total number of {{semantic5:status}} status orders? Create a JSON with keys: 'high_value_orders', 'dept_total_value', 'company_unique_products', 'category_avg_amount', 'high_quantity_customers', and 'status_order_count'. Begin by examining the schema to find relevant columns, and then do your analysis. Note that if the requested company data is not present in the database, then assume the answer is 0 for the relevant question."
    scoring_type: "readfile_jsonmatch"
    file_to_read: "{{artifacts}}/{{qs_id}}/business_analysis.json"
    expected_content: "{\"high_value_orders\": {{sqlite_query:SELECT COUNT(*) FROM enterprise_orders o JOIN enterprise_customers c ON o.CUST_REF = c.CUST_ID JOIN enterprise_companies comp ON c.COMP_REF = comp.COMP_ID WHERE comp.COMP_NM = '{{semantic1:company}}' AND c.REGION = '{{semantic2:region}}' AND o.ORD_AMT > {{number1:15000:35000:currency}}:TARGET_FILE[order_db]}}, \"dept_total_value\": {{sqlite_query:SELECT COALESCE(SUM(o.ORD_AMT), 0) FROM enterprise_orders o JOIN enterprise_customers c ON o.CUST_REF = c.CUST_ID WHERE c.DEPT_CD = '{{semantic3:department}}':TARGET_FILE[order_db]}}, \"company_unique_products\": {{sqlite_query:SELECT COUNT(DISTINCT o.PROD_REF) FROM enterprise_orders o JOIN enterprise_customers c ON o.CUST_REF = c.CUST_ID JOIN enterprise_companies comp ON c.COMP_REF = comp.COMP_ID WHERE comp.COMP_NM = '{{semantic1:company}}':TARGET_FILE[order_db]}}, \"category_avg_amount\": {{sqlite_query:SELECT COALESCE(ROUND(AVG(o.ORD_AMT), 2), 0) FROM enterprise_orders o JOIN enterprise_products p ON o.PROD_REF = p.PROD_ID WHERE p.CATEGORY = '{{semantic4:category}}':TARGET_FILE[order_db]}}, \"high_quantity_customers\": {{sqlite_query:SELECT COUNT(DISTINCT c.CUST_ID) FROM enterprise_orders o JOIN enterprise_customers c ON o.CUST_REF = c.CUST_ID WHERE o.QUANTITY > {{number2:70:85}}:TARGET_FILE[order_db]}}, \"status_order_count\": {{sqlite_query:SELECT COUNT(*) FROM enterprise_orders WHERE STAT_CD = '{{semantic5:status}}':TARGET_FILE[order_db]}}}"
    sandbox_setup:
      components:
        - type: "create_sqlite"
          name: "order_db"
          target_file: "{{artifacts}}/{{qs_id}}/{{entity1}}.db"
          content:
            tables:
              - name: "enterprise_companies"
                columns:
                  - {name: "COMP_ID", type: "auto_id"}
                  - {name: "COMP_NM", type: "TEXT", data_type: "company"}
                  - {name: "INDUSTRY", type: "TEXT", data_type: "category"}
                rows: "{{number3:8:15}}"

              - name: "enterprise_products"
                columns:
                  - {name: "PROD_ID", type: "auto_id"}
                  - {name: "PROD_NM", type: "TEXT", data_type: "product"}
                  - {name: "CATEGORY", type: "TEXT", data_type: "category"}
                  - {name: "BASE_PRICE", type: "INTEGER", data_type: "price"}
                rows: "{{number4:60:70}}"

              - name: "enterprise_customers"
                columns:
                  - {name: "CUST_ID", type: "auto_id"}
                  - {name: "CUST_NM", type: "TEXT", data_type: "person_name"}
                  - {name: "COMP_REF", type: "INTEGER", foreign_key: "enterprise_companies.COMP_ID"}
                  - {name: "REGION", type: "TEXT", data_type: "region"}
                  - {name: "DEPT_CD", type: "TEXT", data_type: "department"}
                rows: "{{number5:200:250}}"

              - name: "enterprise_orders"
                columns:
                  - {name: "ORD_ID", type: "auto_id"}
                  - {name: "CUST_REF", type: "INTEGER", foreign_key: "enterprise_customers.CUST_ID"}
                  - {name: "PROD_REF", type: "INTEGER", foreign_key: "enterprise_products.PROD_ID"}
                  - {name: "ORD_AMT", type: "INTEGER", data_type: "currency"}
                  - {name: "QUANTITY", type: "INTEGER", data_type: "score"}
                  - {name: "STAT_CD", type: "TEXT", data_type: "status"}
                rows: "{{number6:750:1000}}"


  #################################################
  #Section 700: DB Processing Failure Mode Analysis

  - question_id: 701
    samples: 30
    template: "How many orders above 20000.00 are there from customers in the {{semantic2:region}} region in {{artifacts}}/{{qs_id}}/{{entity1}}.db? Write a file {{artifacts}}/{{qs_id}}/big_orders_count.txt that contains ONLY the numerical answer, nothing else. Begin by examining the schema to find relevant columns, and then do your analysis."
    scoring_type: "readfile_stringmatch"
    file_to_read: "{{artifacts}}/{{qs_id}}/big_orders_count.txt"
    expected_content: "{{sqlite_query:SELECT COUNT(*) FROM enterprise_orders o JOIN enterprise_customers c ON o.CUST_REF = c.CUST_ID WHERE c.REGION = '{{semantic2:region}}' AND o.ORD_AMT > 20000:TARGET_FILE[order_db]}}"
    sandbox_setup:
      components:
        - type: "create_sqlite"
          name: "order_db"
          target_file: "{{artifacts}}/{{qs_id}}/{{entity1}}.db"
          content:
            tables:
              - name: "enterprise_companies"
                columns:
                  - {name: "COMP_ID", type: "auto_id"}
                  - {name: "COMP_NM", type: "TEXT", data_type: "company"}
                  - {name: "INDUSTRY", type: "TEXT", data_type: "category"}
                rows: "{{number2:8:15}}"

              - name: "enterprise_products"
                columns:
                  - {name: "PROD_ID", type: "auto_id"}
                  - {name: "PROD_NM", type: "TEXT", data_type: "product"}
                  - {name: "CATEGORY", type: "TEXT", data_type: "category"}
                  - {name: "BASE_PRICE", type: "INTEGER", data_type: "price"}
                rows: "{{number3:20:30}}"

              - name: "enterprise_customers"
                columns:
                  - {name: "CUST_ID", type: "auto_id"}
                  - {name: "CUST_NM", type: "TEXT", data_type: "person_name"}
                  - {name: "COMP_REF", type: "INTEGER", foreign_key: "enterprise_companies.COMP_ID"}
                  - {name: "REGION", type: "TEXT", data_type: "region"}
                  - {name: "DEPT", type: "TEXT", data_type: "department"}
                rows: "{{number4:200:250}}"

              - name: "enterprise_orders"
                columns:
                  - {name: "ORD_ID", type: "auto_id"}
                  - {name: "CUST_REF", type: "INTEGER", foreign_key: "enterprise_customers.CUST_ID"}
                  - {name: "PROD_REF", type: "INTEGER", foreign_key: "enterprise_products.PROD_ID"}
                  - {name: "ORD_AMT", type: "INTEGER", data_type: "currency"}
                  - {name: "QUANTITY", type: "INTEGER", data_type: "score"}
                  - {name: "STAT_CD", type: "TEXT", data_type: "status"}
                rows: "{{number5:750:1000}}"

  - question_id: 702
    samples: 30
    template: "How many orders above 20000.00 are there from customers in the {{semantic2:region}} region in {{artifacts}}/{{qs_id}}/{{entity1}}.db? Begin by examining the schema to find relevant columns, and then do your analysis. When providing your final answer, reply in JSON format using 'num_big_orders' as key - JSON only, no additional text."
    scoring_type: "jsonmatch"
    expected_response: "{\"num_big_orders\": {{sqlite_query:SELECT COUNT(*) FROM enterprise_orders o JOIN enterprise_customers c ON o.CUST_REF = c.CUST_ID WHERE c.REGION = '{{semantic2:region}}' AND o.ORD_AMT > 20000:TARGET_FILE[order_db]}}}"
    sandbox_setup:
      components:
        - type: "create_sqlite"
          name: "order_db"
          target_file: "{{artifacts}}/{{qs_id}}/{{entity1}}.db"
          content:
            tables:
              - name: "enterprise_companies"
                columns:
                  - {name: "COMP_ID", type: "auto_id"}
                  - {name: "COMP_NM", type: "TEXT", data_type: "company"}
                  - {name: "INDUSTRY", type: "TEXT", data_type: "category"}
                rows: "{{number2:8:15}}"

              - name: "enterprise_products"
                columns:
                  - {name: "PROD_ID", type: "auto_id"}
                  - {name: "PROD_NM", type: "TEXT", data_type: "product"}
                  - {name: "CATEGORY", type: "TEXT", data_type: "category"}
                  - {name: "BASE_PRICE", type: "INTEGER", data_type: "price"}
                rows: "{{number3:20:30}}"

              - name: "enterprise_customers"
                columns:
                  - {name: "CUST_ID", type: "auto_id"}
                  - {name: "CUST_NM", type: "TEXT", data_type: "person_name"}
                  - {name: "COMP_REF", type: "INTEGER", foreign_key: "enterprise_companies.COMP_ID"}
                  - {name: "REGION", type: "TEXT", data_type: "region"}
                  - {name: "DEPT", type: "TEXT", data_type: "department"}
                rows: "{{number4:200:250}}"

              - name: "enterprise_orders"
                columns:
                  - {name: "ORD_ID", type: "auto_id"}
                  - {name: "CUST_REF", type: "INTEGER", foreign_key: "enterprise_customers.CUST_ID"}
                  - {name: "PROD_REF", type: "INTEGER", foreign_key: "enterprise_products.PROD_ID"}
                  - {name: "ORD_AMT", type: "INTEGER", data_type: "currency"}
                  - {name: "QUANTITY", type: "INTEGER", data_type: "score"}
                  - {name: "STAT_CD", type: "TEXT", data_type: "status"}
                rows: "{{number5:750:1000}}"

  - question_id: 703
    samples: 30
    template: "How many orders above 20000.00 are there from customers in the {{semantic2:region}} region in {{artifacts}}/{{qs_id}}/{{entity1}}.db? Begin by examining the schema to find relevant columns, and then do your analysis. When providing your final answer, provide only the numerical answer with no explanation, punctuation, or additional text."
    scoring_type: "stringmatch"
    expected_response: "{{sqlite_query:SELECT COUNT(*) FROM enterprise_orders o JOIN enterprise_customers c ON o.CUST_REF = c.CUST_ID WHERE c.REGION = '{{semantic2:region}}' AND o.ORD_AMT > 20000:TARGET_FILE[order_db]}}"
    sandbox_setup:
      components:
        - type: "create_sqlite"
          name: "order_db"
          target_file: "{{artifacts}}/{{qs_id}}/{{entity1}}.db"
          content:
            tables:
              - name: "enterprise_companies"
                columns:
                  - {name: "COMP_ID", type: "auto_id"}
                  - {name: "COMP_NM", type: "TEXT", data_type: "company"}
                  - {name: "INDUSTRY", type: "TEXT", data_type: "category"}
                rows: "{{number2:8:15}}"

              - name: "enterprise_products"
                columns:
                  - {name: "PROD_ID", type: "auto_id"}
                  - {name: "PROD_NM", type: "TEXT", data_type: "product"}
                  - {name: "CATEGORY", type: "TEXT", data_type: "category"}
                  - {name: "BASE_PRICE", type: "INTEGER", data_type: "price"}
                rows: "{{number3:20:30}}"

              - name: "enterprise_customers"
                columns:
                  - {name: "CUST_ID", type: "auto_id"}
                  - {name: "CUST_NM", type: "TEXT", data_type: "person_name"}
                  - {name: "COMP_REF", type: "INTEGER", foreign_key: "enterprise_companies.COMP_ID"}
                  - {name: "REGION", type: "TEXT", data_type: "region"}
                  - {name: "DEPT", type: "TEXT", data_type: "department"}
                rows: "{{number4:200:250}}"

              - name: "enterprise_orders"
                columns:
                  - {name: "ORD_ID", type: "auto_id"}
                  - {name: "CUST_REF", type: "INTEGER", foreign_key: "enterprise_customers.CUST_ID"}
                  - {name: "PROD_REF", type: "INTEGER", foreign_key: "enterprise_products.PROD_ID"}
                  - {name: "ORD_AMT", type: "INTEGER", data_type: "currency"}
                  - {name: "QUANTITY", type: "INTEGER", data_type: "score"}
                  - {name: "STAT_CD", type: "TEXT", data_type: "status"}
                rows: "{{number5:750:1000}}"
\end{lstlisting}

\end{document}